\documentclass[10pt,twocolumn,letterpaper]{article}

\usepackage{cvpr}              

\usepackage{graphicx}
\usepackage{amsmath}
\usepackage{amssymb}
\usepackage{booktabs}
\usepackage{overpic}
\usepackage[linesnumbered,ruled]{algorithm2e}
\usepackage{listings}
\usepackage{makecell}

\usepackage{pifont}
\newcommand{\mathvec}[1]{\boldsymbol{#1}}
\newcommand{\mathmat}[1]{\mathbf{#1}}
\newcommand{\mathset}[1]{\mathcal{#1}}

\newcommand{\figref}[1]{Fig.~\ref{#1}}
\newcommand{\tabref}[1]{Tab.~\ref{#1}}
\newcommand{\eqnref}[1]{Eqn.~(\ref{#1})}

\graphicspath{{./figures/}}
\newcommand{\change}[1]{\textcolor{black}{#1}}

\def\eg{\emph{e.g.}}

\def\improveb#1{{\footnotesize  \color[rgb]{0.27, 0.71, 0.45} (+#1)}}
\def\decrease#1{{\footnotesize  \color[rgb]{0.85, 0.15, 0.15} (-#1)}}

\definecolor{cvprblue}{rgb}{0.21,0.49,0.74}
\usepackage[pagebackref,breaklinks,colorlinks,citecolor=cvprblue]{hyperref}

\def\paperID{16183} 
\def\confName{CVPR}
\def\confYear{2025}

\title{VoxelSplat: Dynamic Gaussian Splatting as an Effective Loss for Occupancy and Flow Prediction}

\author{Ziyue Zhu\textsuperscript{1} \quad 
        Shenlong Wang\textsuperscript{2} \quad 
        Jin Xie\textsuperscript{3}\textsuperscript{\textdagger} \quad 
        Jiang-jiang Liu\textsuperscript{4} \quad 
        Jingdong Wang\textsuperscript{4} \quad 
        Jian Yang\textsuperscript{1}\textsuperscript{\textdagger}\\
        \textsuperscript{1}PCA Lab, VCIP, College of Computer Science, Nankai University \quad
        \textsuperscript{2}UIUC \quad \\
        \textsuperscript{3}School of Intelligence Science and Technology, Nanjing University, Suzhou \quad
        \textsuperscript{4}Baidu \\
        {\tt\small zhuziyue@mail.nankai.edu.cn \quad 
        csjxie@nju.edu.cn \quad
        csjyang@nankai.edu.cn}
}

\begin{document}
\maketitle
\footnotetext[1]{\textsuperscript{\textdagger} Corresponding authors: Jian Yang and Jin Xie.}
\begin{abstract}
    
Recent advancements in camera-based occupancy prediction have focused on the simultaneous prediction of 3D semantics and scene flow, a task that presents significant challenges due to specific difficulties, e.g., occlusions and unbalanced dynamic environments. In this paper, we analyze these challenges and their underlying causes.
To address them, we propose a novel regularization framework called VoxelSplat. This framework leverages recent developments in 3D Gaussian Splatting to enhance model performance in two key ways:
\(\left( i \right)\) Enhanced Semantics Supervision through 2D Projection: During training, our method decodes sparse semantic 3D Gaussians from 3D representations and projects them onto the 2D camera view. This provides additional supervision signals in the camera-visible space, allowing 2D labels to improve the learning of 3D semantics.
\(\left( ii \right)\) Scene Flow Learning: Our framework uses the predicted scene flow to model the motion of Gaussians, and is thus able to learn the scene flow of moving objects in a self-supervised manner using the labels of adjacent frames.
Our method can be seamlessly integrated into various existing occupancy models, enhancing performance without increasing inference time. Extensive experiments on benchmark datasets demonstrate the effectiveness of VoxelSplat in improving the accuracy of both semantic occupancy and scene flow estimation.
The project page and codes are available at \href{https://zzy816.github.io/VoxelSplat-Demo/}{https://zzy816.github.io/VoxelSplat-Demo/}.
\end{abstract}
\section{Introduction}
\label{sec:intro}

Robust and accurate perception is crucial for self-driving systems. 
Camera-centric occupancy map perception has become popular in both industry
and academia~\cite{huang2023tri, li2023fb, huang2021bevdet, sima2023_occnet, li2022bevformer, pan2023renderocc, wang2023openoccupancy, zhang2023occformer, li2023voxformer, huang2023selfocc} due to its low-cost sensors, robustness, generalizability, and seamless integration with motion planning. 
Joint semantic and flow prediction on occupancy maps shows great potential, 
as it can handle both semantic and dynamic understanding, which are key elements for safe driving.

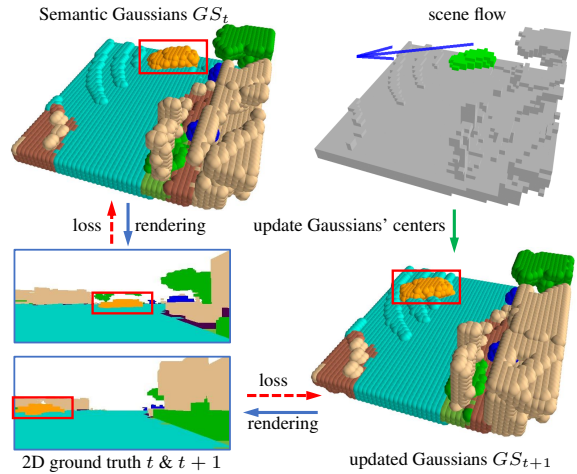
\begin{figure}[t!]
    \centering
    \begin{overpic}[width=0.9\linewidth]{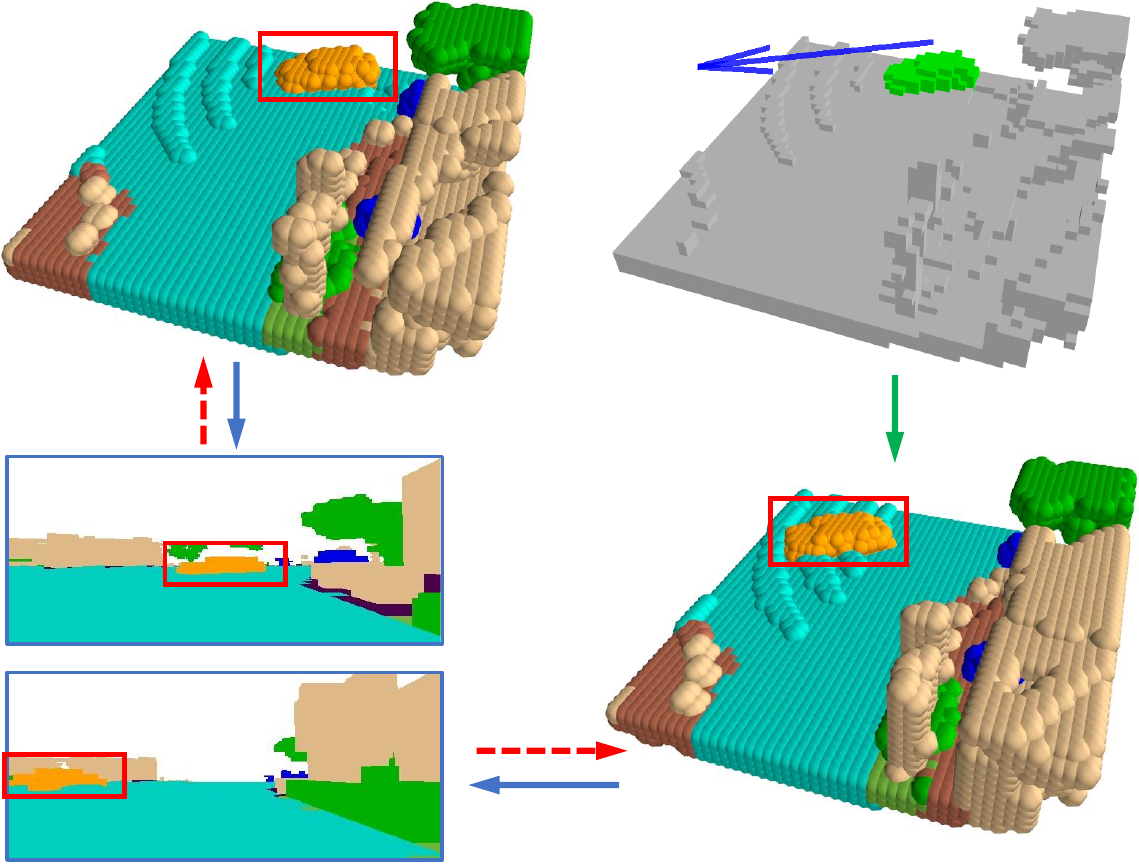}
    \put(5, 76.5){\scriptsize{Semantic Gaussians $GS_{t}$}}
    \put(74, 76.5){\scriptsize{scene flow}}
    \put(60, -2.4){\scriptsize{updated Gaussians $GS_{t+1}$}}
    \put(2, -3){\scriptsize{2D ground truth $t$ \& $t+1$}}
    \put(43, 39){\scriptsize{update Gaussians' centers}}
    \put(44, 11.5){\scriptsize{loss}}
    \put(42, 4){\scriptsize{rendering}}
    \put(11, 39){\scriptsize{loss}}
    \put(22, 39){\scriptsize{rendering}}
    \end{overpic}
    \vspace{4pt}
    \caption{During training, our method additionally predicts 3D Gaussians $GS_{t}$ representing semantic logits of occupied regions in the current frame. We then obtain Gaussians $GS_{t+1}$ for the future frame by updating their centers using the predicted scene flow. By rendering these Gaussians into the camera views of different time stamps, both semantic and scene flow predictions can be supervised using the multi-frame 2D ground truths.
}
    \label{fig:ts}
\end{figure}

Despite advancements, camera-based occupancy perception faces key challenges. 
\change{First, large portions of the annotated regions \cite{tian2024occ3d, wang2023openoccupancy, sima2023_occnet} are inaccessible to the camera views due to \textbf{occlusion}. The supervision intended to improve performance on these invisible regions may propagate misleading signals along the camera rays to the image features, potentially causing adverse effects and deteriorating performance rather than enhancing it.
}
Second, voxelized representations, 
while advantageous for their grid structure and planner integration, struggle with {\bf explicit motion modeling}. 
Voxels are suited for implicit Eulerian motion, 
yet explicit Laplacian motion is more critical for safe planning in self-driving. 
Finally, most benchmarks and datasets \cite{caesar2020nuscenes, wang2023openoccupancy} suffer from {\bf class \change{and speed} imbalances}, 
especially for high-speed objects, 
limiting robustness to rare dynamic events and hindering safe driving progress. 

To address these challenges, we present VoxelSplat, a novel occupancy perception framework. At the core of our method is a new Gaussian splatting-based training mechanism. Inspired by the recent success of Gaussian splatting in 3D scene modeling and novel view rendering, we incorporate dynamic Gaussians into our voxelized representation as an additional header, enabling our voxels to render semantics and motion to various viewpoints as images, and calculate and minimize rendering losses. 
Thanks to the explicit nature of Gaussians, we can {\it explicitly model motion} as dynamic movement on Gaussian points, rendering them onto virtual image views to receive supervision through differentiable splat rendering. 
\change{As shown in \figref{fig:ts}, 
our method models the occupancy semantic field with 3D Gaussians and their motion with predicted scene flows. 
The semantic field of the next frame is predicted by transforming the Gaussians using the flows. 
2D ground truths from adjacent frames provide supervision for both semantics and flows.
}
Importantly, {\it all additional Gaussian splat rendering and supervision occur only during training} as extra rendering headers and losses. During inference, we maintain the original voxelized pipeline without any additional overhead, while benefiting from increased accuracy and robustness.

We evaluate VoxelSplat across various benchmarks, showing improvements over state-of-the-art methods in both semantic accuracy by 3.6\% and flow estimation accuracy by 20.2\%. Additionally, we demonstrate the flexibility of VoxelSplat as a training-time plugin that enhances performance across diverse occupancy prediction architectures. 
We will release the code upon acceptance, 
providing the community with an effective tool to boost the training of occupancy perception networks.
\change{In summary, our contributions include:
\begin{itemize}
  \itemsep0em
  \item We propose a plug-and-play loss framework that utilizes dynamic Gaussian splatting to boost the learning of both occupancy and flow prediction.
  \item Extensive experiments on benchmark datasets demonstrate that our proposed framework significantly improves the performance of both semantic and flow prediction across various occupancy architectures.
\end{itemize}} 
\section{Related Work}
\subsection{Camera-based Occupancy Prediction}
Occupancy prediction \cite{gan2024gaussianocc, song2024collaborative, ma2024cam4docc, wang2023panoocc, ma2024cotr, peng2024learning, zhao2024lowrankocc, tang2024sparseocc, vobecky2024pop, zhang2023occnerf, zheng2025occworld, wu2024deep} has demonstrated significant advantages in 3D scene understanding, making it a crucial task in autonomous driving research \cite{huang2023tri, huang2023selfocc, li2022bevformer, li2023fb,  zhang2023occformer, tian2024occ3d, miao2023occdepth, gan2024gaussianocc, yu2023flashocc, yan2022rignet, yan2024tri}. 
Unlike traditional object detection paradigms, occupancy perception offers several benefits: it expresses dense 3D geometry, accurately provides spatial locations for objects beyond predefined categories, and describes the shapes of irregular obstacles. These advantages have led to a surge in research focused on occupancy prediction tasks \cite{tian2024occ3d, wang2023openoccupancy, tong2023scene}, which predict the occupancy status in the region of interest around the ego vehicle from point clouds or images.
Recent methods typically divide the space into voxel grids, estimating the occupancy status and semantics of each grid. For example, 
BEVDet4D \cite{huang2021bevdet} directly predict the occupancy from bev features.
SurroundOcc \cite{wei2023surroundocc} proposed a surround-view 3D occupancy perception method that uses spatial 2D-3D attention to lift image features into 3D space, and designed a pipeline to convert point clouds to dense occupancy ground truth. Huang \etal \cite{huang2023tri} employ the representation of tri-plane to represent the occupancy field.
Similarly, VoxFormer \cite{li2023voxformer} employed an depth-based approach for camera-based semantic occupancy prediction. FB-OCC \cite{li2023fb} introduced a novel forward-backward projection method to address the insufficient BEV feature density of forward projection and the mismatches in 2D and 3D space caused by backward projection. 
In addition to end-to-end supervision of the 3D grid's semantics, there are other supervision methods as well. 
Surroundsdf \cite{liu2024surroundsdf} implicitly predicts
the signed distance field (SDF) and semantic field for the
continuous perception from surround images.
RenderOcc \cite{pan2023renderocc} predict a neural radiance field and use 2D labels as supervision.
Despite these advancements, these methods are limited by the modeling to dynamic objects.

\subsection{3D Gaussian Splatting}

The recent groundbreaking work \cite{3dgs, tang2023dreamgaussian, feng2024new, lyu20243dgsr, yi2023gaussiandreamer, yan2024street, zhao2024tclc, wei2024omniscene} represents static scenes using Gaussians, with positions and appearance learned via a differentiable splatting-based renderer. Notably, 3D Gaussian Splatting (3DGS) \cite{3dgs, chen2024omnire, blinn1982generalization} delivers impressive real-time rendering performance through Gaussian split/clone operations and an efficient splatting-based rendering technique. 
Furthermore, the recent advanced works explore the potential of Dynamic 3D Gaussians \cite{lu20243d, yang2023deformable, luiten2023dynamic, bae2024per, chu2024dreamscene4d, liu2024modgs, zhu2024motiongs} in modeling dynamic scenes by representing objects and their movements through time-conditioned Gaussian distributions in a 3D space. 
Deformable3DGS \cite{yang2023deformable} learns temporal motion and rotation of each 3D Gaussian, making it ideal for dynamic tracking. Similarly, \cite{luiten2023dynamic} predict temporal movements of 3DGS. 
RealTime4DGS \cite{duan20244d} employs 4D Gaussian representation for 3D dynamics but uses a 4D rotation formulation, which is less interpretable and lacks spatial-temporal separability compared to rotor-based representation. 

In this work,
we explore the potential of dynamic Gaussians \cite{gao2024gaussianflow, guo2024motion, labe2024dgd, lee2024fully, zhang2024egogaussian, katsumata2025compact, huang2024sc, lin2024gaussian} for autonomous driving perception
by proposing a loss where driving scenes are represented by dynamic Gaussians.
We demonstrate that compared to typical BEV and NeRF representations in perception,
dynamic Gaussians better capture geometry and motion, enhancing model performance.



\begin{figure*}[ht!]
    \centering
    \begin{overpic}[width=0.97\linewidth]{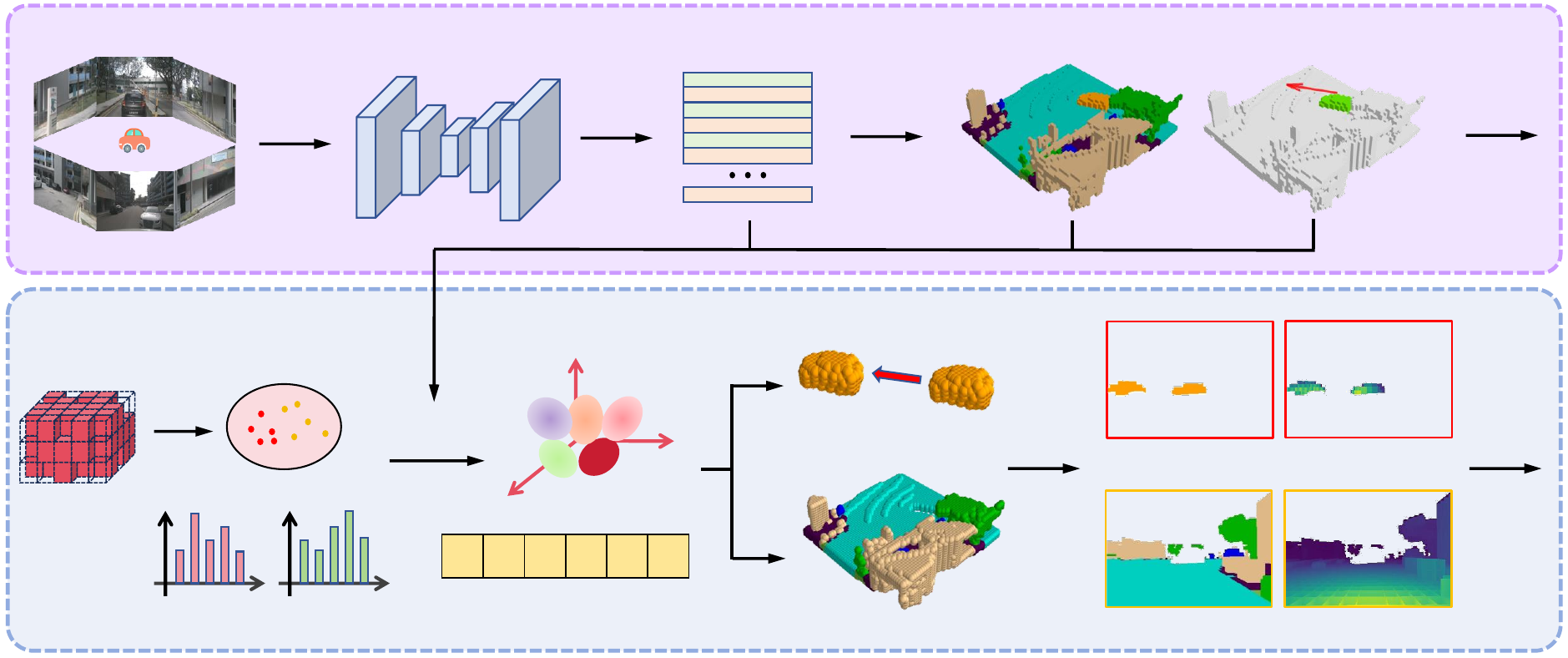}
        \put(1, 26){\scriptsize{multi-view images sequence}}
        \put(24, 39){\scriptsize{occupancy network}}
        \put(42, 39){\scriptsize{voxel embeddings}}
        \put(54.5, 34){\scriptsize{decode}}
        \put(65, 39){\scriptsize{semantics}}
        \put(81, 39){\scriptsize{scene flow}}
        \put(94, 34.2){\textbf{\scriptsize{$\mathcal{L}_{3D}$}}}
        \put(94, 13){\textbf{\scriptsize{$\mathcal{L}_{2D}$}}}
        \put(1, 9.5){\scriptsize{ground truth}}
        \put(13, 18.5){\scriptsize{weighted sampling}}
        \put(12, 2.5){\scriptsize{class}}
        \put(19.5, 2.5){\scriptsize{speed}}
        \put(29.5, 2.5){\scriptsize{dynamic 3D gaussians}}
        \put(50, 21){\scriptsize{update centers: $\mu+\Delta x$}}
        \put(64, 13){\scriptsize{rendering}}
        \put(23, 14){\scriptsize{gather \& decode}}
        \put(2, 21){\textbf{training only}}
        \put(2, 39){\textbf{training \& inference}}
        \put(71, 12.3){\scriptsize{dynamic semantics depth $t \& t+1$}}
        \put(74, 1.7){\scriptsize{static semantics depth $t$}}
        \put(29, 6){\small{$\mu$}}
        \put(31, 6){\scriptsize{$\Delta x$}}
        \put(34.5, 6){\small{$s$}}
        \put(36.8, 6){\small{$\alpha$}}
        \put(39.5, 6){\small{$r$}}
        \put(42, 6){\small{$s_c$}}
    \end{overpic}
    \caption{The overview of our framework: 
(1) Employing an occupancy model integrated with our flow decoder to predict occupancy and scene flow.
(2) Sampling coordinates from occupied voxel centers using ground truth labels to extract features, semantic logits, and scene flow. Then, 3D semantic Gaussians are decoded.
(3) Dividing the Gaussians into static and dynamic types, with dynamic ones updated by predicted scene flow.
(4) Rendering static and dynamic Gaussians separately for 2D supervision. 
}
    \label{fig:framework}
\end{figure*}
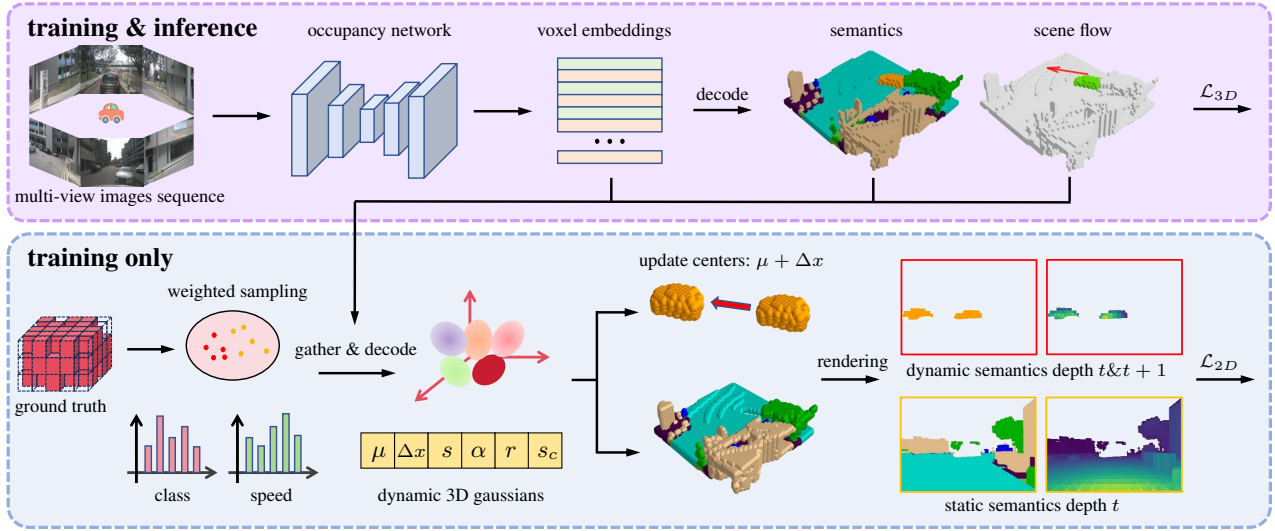
\label{sec:method}
\section{Method}

In this section, we first revisit the concept of 3D Gaussian Splatting as a preliminary (\ref{sec:pre}). 
We then elaborate on the technical details of our framework, 
including our strategy for predicting 3D Gaussians, decomposing dynamic and static objects, and the forward rendering process (\ref{sec:architecture}).
Finally, we introduce how our rendering regularization losses (\ref{sec:splatgs})
enhance the learning of both 3D semantics and scene flow.
An overview of our framework is shown in \figref{fig:framework}.

\subsection{Preliminary: Semantic Gaussian Splatting} 
\label{sec:pre}



\paragraph{3D Gaussians}
3D Gaussian Splatting (3DGS) \cite{3dgs} has demonstrated the capability to achieve real-time, state-of-the-art rendering quality for complex scenes. This technique encodes a scene as a dense collection of \(N\) anisotropic 3D Gaussian ellipsoids, where each Gaussian is fully characterized by its 3D covariance matrix \(\mathmat{\Sigma}\) and its center position \(\mathvec{\mu}\):  
\begin{equation}
G(\mathvec{x}) = \exp\left( -\frac{1}{2} (\mathvec{x} - \mathvec{\mu})^T \mathmat{\Sigma}^{-1} (\mathvec{x} - \mathvec{\mu}) \right).
\end{equation}  
Here, $\mathmat{\Sigma} = \mathmat{R} \mathmat{S_c} \mathmat{S}_c^T \mathmat{R}^T$ and \(\mathmat{S_c} = \text{diag}(s_x, s_y, s_z) \in \mathbb{R}^3\) denotes the anisotropic scaling factors, and \(\mathmat{R} \in SO(3)\) is the rotation matrix, parameterized as a quaternion. Both \(\mathmat{S}\) and \(\mathmat{R}\) are treated as learnable parameters. 

In addition to \(\mathvec{\mu}\), \(\mathbf{S}\), and \(\mathbf{R}\), each Gaussian is further associated with an opacity parameter \(\alpha \in (0, 1)\), which governs the transparency of the Gaussian. 
Compared with NeRF and dense BEV features,
3D Gaussians provide a more explicit representation of the 3D scene, 
effectively capturing object surfaces.
Hence, this representation can be utilized for occupancy prediction by modeling the occupied regions.

\paragraph{Gaussian Splatting for Semantics.}
To render RGB images, spherical harmonic (SH) coefficients in \(\mathbb{R}^k\) are employed for each Gaussian to encode view-dependent color information, where \(k\) is determined by the SH order. However, in tasks such as occupancy and flow prediction, color information is unnecessary. Therefore, we replace the SH coefficients with semantic logits. 
The blending process is governed by the following equations:
\begin{equation}\label{eq:render}
    S = \sum_{i=1}^M s_i \alpha_i \prod_{j=1}^{i-1} (1 - \alpha_j), \hspace{2pt}
    D = \sum_{i=1}^M d_i \alpha_i \prod_{j=1}^{i-1} (1 - \alpha_j),
\end{equation} 
where \(S\) represents the accumulated semantic logits, and \(D\) represents the depth accumulation for proper depth-aware rendering. 
In the lower right corner of \figref{fig:framework}, 
we render the semantics and depth for supervision.


\subsection{VoxelSplat Architecture}
\label{sec:architecture}

As shown in \figref{fig:framework}, 
we use a backbone voxel architecture to predict voxel features, semantic logits, and scene flows. 
Weighted point sampling selects Gaussian centers, from which dynamic Gaussians are decoded. The Gaussians are assigned logits and flows to model semantics and motion. 
Finally, we splat the Gaussians into camera views for supervision.


\noindent\textbf{Backbone Voxel Architecture.}
Starting from the occupancy network architectures \cite{huang2021bevdet, li2022bevformer, wang2023panoocc, wang2023openoccupancy, tian2024occ3d}, we input multiple consecutive frames of multi-view images to the model. This allows the model to leverage temporal information, providing a more detailed understanding of the dynamic driving scene.
The model decodes a voxel feature $\mathvec{V}$, which captures the temporal aspects of the driving environment.
In addition to decoding the voxel feature $\mathvec{V}$, 
the model also predicts 3D semantics $\mathmat{S}$ and scene flow $\mathmat{F}$. 
The 3D semantics $\mathmat{S}$ involves classifying different regions within the scene into categories. 
Scene flow $\mathmat{F}$ represents the motion of objects within the scene over time, providing insights into the dynamics of the driving environment. 
Both the 3D semantics $\mathmat{S}$ and scene flow $\mathmat{F}$ predictions can be supervised using the 3D annotations \cite{liu2023fully}.

\noindent\textbf{Weighted Points Sampling.}
After obtaining the voxel features \(\mathvec{V}\), scene flow \(\mathvec{F}\), and 3D semantics \(\mathmat{S}\), we additionally predict 3D Gaussians, aiming to project and supervise the scene flow and semantics in the camera view. Initially, using the ray casting toolbox \cite{liu2023fully}, we generate camera masks that indicate the visible regions of the scene. From the occupied grids within these visible regions, we sample a set of voxel center points to serve as the centers of the Gaussians.

To address the issue of class imbalance and varying speed distributions, we design a simple yet effective algorithm to balance the sampling process. For each data batch containing \(P\) semantic types, we further divide the voxels into \(Q\) classes based on their speed, ranging from slow to fast. This results in a total of \(PQ\) classes. For each class \((p, q)\), which contains \(N_{p,q}\) voxels, the probability of sampling voxels from this class is computed as:
\begin{equation}\label{eq:probability}
    p_{p,q} = \frac{P(N_{p,q})}{\sum_{i=1}^{PQ} P(N_{p,q})}, \hspace{10pt}
    P(x) = \frac{1}{x^t + 1},
\end{equation} 
where \(P(x)\) is a function that mitigates the effect of class size imbalance by scaling the probabilities.

Through this process, we obtain a set of 3D coordinates \(\{ \mathvec{\mu}_n \}_{n=1}^{N}\).
Using these points, we apply a grid sampling strategy to query the output from the occupancy network, 
gathering the corresponding voxel embeddings \(\{ \mathvec{v}_n \}_{n=1}^{N}\), scene flows\(\{ \Delta \mathvec{x}_n \}_{n=1}^{N}\), and semantic logits \(\{ \mathvec{s}_n \}_{n=1}^{N}\).
Note that $\Delta \mathvec{x}$  is obtained by multiplying the original predicted scene flow by the time interval between two frames, 
so it represents the movement vector of the object.

\noindent\textbf{Decode Dynamic Gaussians.}
Then, we decode the gathered coordinates, embeddings, flow and semantics \(\{ (\mathvec{\mu}_n, \mathvec{s}_n, \Delta \mathvec{x}_n, \mathvec{v}_n) \}_{n=1}^{N}\) into 4D gaussians.
Specifically, 
we directly use the \(\{ \mathvec{\mu}_n \}_{n=1}^{N}\) and \(\{ \mathvec{s}_n \}_{n=1}^{N}\) to represent the centers and semantics logits of Gaussians,
while $\{ \Delta \mathvec{x}_n \}_{n=1}^{N}$ to denote the movement of Gaussians' centers from the current frame to the next frame.
Furthermore,
we employ a simple two layers MLP as Gaussians Decoder $G(\mathvec{x})$ to
decode the voxel embeddings $\{ \mathvec{v}_n \}_{n=1}^{N}$, 
which contain the information of 3D scenes, 
into the shape attributes of Gaussians, including opacity $\mathvec{\alpha}$, rotation $\mathvec{r}$, and scaling $\mathvec{s}_c$. 
Given the relatively low resolution of the voxel space, 
we add learnable positional embeddings $\mathvec{pe}$ to the embeddings $\{ \mathvec{v}_n \}_{n=1}^{N}$. 
The equations are as follows:
\begin{equation}
\begin{aligned}
    g_n : (\mathvec{\mu}, \Delta \mathvec{x}, \mathvec{s}, \mathvec{\alpha}, \mathvec{r}, \mathvec{s}_c) &= (\mathvec{\mu}_n, \Delta \mathvec{x}_n, \mathvec{s}_n, G(\mathvec{v}_n+ \mathvec{pe}) ). \\
\end{aligned}
\end{equation}

In this way, we use $\mathset{G}=\{ \mathvec{g}_n \}_{n=1}^{N}$ to denote the Gaussians.
For better learning the scene flow of the moving objects,
we decompose the Gaussians into static $\mathset{G}^s=\{ \mathvec{g}_s \}_{s=1}^{S}$ and dynamic ones $\mathset{G}^d=\{ \mathvec{g}_d \}_{d=1}^{D}$, 
according to the semantics of the Gaussians' centers in the ground truth.
With the corresponding estimated scene flow \(\{ \Delta \mathvec{x}_n \}_{d=1}^{D}\),
we update the gaussians’ centers with $\mathvec{\mu} + \Delta \mathvec{x}$ and
predict the dynamic gaussians of the future frame $\mathset{G}^{df}=\{ \mathvec{g}^f_d \}_{d=1}^{D}$.

\noindent\textbf{Splatting Rendering the Current and Future Gaussians.}
After obtaining static Gaussians $\mathset{G}^{s}$, 
and dynamic ones of current $\mathset{G}^{d}$ and future frame $\mathset{G}^{df}$, 
we apply the fast differentiable Gaussian rasterization method \cite{3dgs} to render the 2D depth maps and semantic maps.

Using \eqnref{eq:render}, 
we first splat the static Gaussians $\mathset{G}^{s}$ onto the 2D camera plane and render the semantic maps and the depth maps $(\mathmat{S}^s, \mathmat{D}^s)$. 
Subsequently, we splat both the future dynamic Gaussians of current frame and the future frame together $\mathset{G}^{d} \cup \mathset{G}^{df}$ and obtain $(\mathmat{S}^d, \mathmat{D}^d)$ for unsupervised scene flow learning. 


\subsection{Training and inference}\label{sec:splatgs}

To supervise our predictions, 
we employ virtual view rendering to boost training, a simple online strategy to generate 2D ground truth, 
and a joint loss to optimize predictions.


\paragraph{Virtual View Rendering.}
In addition to rendering the Gaussians from the current frame's camera views, 
we also select views from adjacent frames. 
This strategy provides more multi-view cues, 
aiding the occupancy model in emphasizing future location perception. 
Finally, we obtain \(M\) rendering views for static objects \(\{(\mathmat{S}_k^s, \mathmat{D}_k^s)\}_{k=1}^{M}\) and dynamic objects \(\{(\mathmat{S}_k^d, \mathmat{D}_k^d)\}_{k=1}^{M}\).



\noindent\textbf{Online Label Generation.}
In line with the way 2D predictions are generated, 
we separate the ground truth voxels into static and dynamic ones.
The dynamic ones are duplicated and moved according to the scene flow.
With the the efficient 3DGS tools \cite{3dgs}, 
2D labels are generated by projecting the static and dynamic voxels into camera views. 
In this way semantics and depth labels of static objects $(\hat{\mathmat{S}}^s, \hat{\mathmat{D}}^s)$ and dynamic ones $(\hat{\mathmat{S}}^d, \hat{\mathmat{D}}^d)$ are obtained.

\noindent\textbf{Losses of 3D predictions and rendering results.}
As our method is built upon existing methods \cite{li2022bevformer, huang2021bevdet, li2023fb, liu2023fully},
we use the original occupancy loss function $\mathcal{L}_{occ}$ of these methods to supervise the semantics.
For scene flow prediction, we apply L\_1 loss and assign weights to the voxels based on the magnitude of their speed.
Our 3D loss $\mathcal{L}_{3D}$ is the combination of occupancy loss and scene flow loss.

To supervise the rendering results of the 4D gaussians,
we employ cross-entropy (CE) loss for the semantic prediction supervision and  L\_1 loss for the rendered depth.
Thus, the 2D loss is defined as:
\begin{equation}
    \mathcal{L}_{2D} = \sum_{k=p, q} \sum_{m=1}^M CE(\mathmat{S}_m^k, \hat{\mathmat{S}}_m^k) + ||\mathmat{D}_m^k - \hat{\mathmat{D}}_m^k||_1 \hspace{10pt}. 
\end{equation}
Note that our 2D loss is only calculated on areas with rendered values.
Finally, the total loss $\mathcal{L}_{total}$ is the sum of 2D loss  and 3D loss .

\noindent\textbf{Inference.}
As our VoxelSplat framework is a plug-and-play training mechanism to boost performance, 
the rendering process is not needed in inference. 
We only employ the voxel-based pipeline, as denoted by the upper branch in \figref{fig:framework}. 
Thus, our framework provides an effective loss design with no additional cost during inference.









\section{Experiments}

\subsection{Settings}

\noindent\textbf{Dataset and Metrics.}
We train and evaluate our model on the nuScenes dataset \cite{caesar2020nuscenes}, which consists of large-scale multimodal data collected from 6 surround-view cameras, 1 LiDAR sensor, and 5 radar sensors. The dataset contains 1000 video sequences, and is divided into 700/150/150 splits for training, validation, and testing, respectively. 
The annotations for occupancy and flow ground truth are provided by OpenOcc \cite{sima2023_occnet, liu2023fully}, which is used as the benchmark in the CVPR 2024 workshop challenge. 
OpenOcc partitions each key-frame scene in the nuScenes into $H \times W \times D$ grids, providing 3D ground truth for semantic occupancy ($H \times W \times D$) and x-y-direction scene flow ($H \times W \times D \times 2$).
Additionally, to ensure a comprehensive comparison across a wide range of methods, we also train our models using the annotations provided by Occ3D \cite{tian2024occ3d} and SurroundOcc \cite{wei2023surroundocc}.

Following the recent work \cite{liu2023fully, li2023fb, huang2024gaussian}, we evaluate our occupancy prediction using the RayIoU and mIoU metrics. Additionally, we evaluate the quality of predicted scene flow by measuring the velocity error for a set of true positives (TP) using a 2-meter distance threshold. The absolute velocity error (AVE) is calculated for 8 dynamic classes 

\noindent\textbf{Implementation details.} 
Since our method is to introduce flow prediction and loss functions on top of existing models,
we conduct our experiments based on three advanced models \cite{huang2021bevdet, li2023fb, liu2023fully} for occupancy prediction. 
As there are no publicly available open-source models for occupancy and flow prediction, 
we modify the baseline by adding two linear layers to the decoder of the occupancy models. 
During inference,
we only predict the flow belonging to the dynamic object classes.
The learning rate, optimization strategies, and input image size remain consistent with the original settings of the respective models. 

In the the process of our Weighted Points Sampling strategy,
the scene flow are separated to 6 categories in \figref{fig:ratio} according to the speed and we totally sample 100000 points for each scene in a batch.

\subsection{Qualitative and Quantitative Comparison}
In this section, we employ RayIoU and mIoU to evaluate the quality of the predicted occupancy, and mAVE to assess the accuracy of the predicted scene flow.
Meanwhile, we visualize our prediction results of both occupancy and scene flow. The ground truth and results of other methods are also provided for comparison.
 
\noindent\textbf{Quantitative Results.} 
In \tabref{tab:performance} and \tabref{tab:nuscseg}, our VoxelSplat model is built by integrating our flow decoder and rendering loss designs into FB-Occ \cite{li2023fb}.

In \tabref{tab:performance}, we compare the quantitative performance of our method with several state-of-the-art occupancy prediction models \cite{liu2023fully, li2023fb, huang2021bevdet, pan2023renderocc, li2022bevformer}. The results show that by adding our lightweight scene flow decoder, the occupancy models \cite{huang2021bevdet, li2023fb} are able to predict the scene flow successfully, without a significant drop in semantic prediction performance. 
Our VoxelSplat outperforms previous methods across all metrics. Compared to FB-Occ, our VoxelSplat achieves an improvement of 3.4 and 3.1 in occupancy prediction, measured by RayIoU and mIoU, respectively. For scene flow prediction, VoxelSplat provides a performance improvement of 0.202 in mAVE.

In \tabref{tab:nuscseg}, we train our model using the OpenOccupancy annotations and present the mIoU results for detailed categories. Since only semantic annotations are provided, our flow decoder and speed-based weighted sampling are not employed in this experiment. 
Overall, our VoxelSplat achieves improvements of 3.48 and 3.96 in occupancy prediction (measured by IoU and mIoU) compared to FB-Occ \cite{li2023fb}. Specifically, our method demonstrates significant improvements in predicting objects (e.g., bicycle, car, pedestrian, and motorcycle) with small proportions, which can be attributed to our weighted sampling strategy.

\begin{table*}[ht]
\centering
\small
\begin{tabular}{l|ccc|ccc|c|c|c}
\toprule
Methods & \multicolumn{3}{c|}{Lidar Occ Flow} & \multicolumn{3}{c|}{RayIoU$_{1m, 2m, 4m}$ $\uparrow$}  & RayIoU $\uparrow$ & mAVE $\downarrow$ & mIoU $\uparrow$\\ 
\midrule
RenderOcc \cite{pan2023renderocc}   & \checkmark & & & 13.4 & 19.6 & 25.5 & 19.5 & - & 24.6 \\
BEVFormer \cite{li2022bevformer}   &  & \checkmark & & 26.1 & 32.3 & 38.0 & 32.4 & - & 39.1 \\
BEVDet-Occ (8f) \cite{huang2021bevdet}   & \checkmark & \checkmark & & 26.0 & 32.4 & 38.2 & 32.0 & - & 39.2 \\
FB-Occ (16f) \cite{li2023fb}    & \checkmark & \checkmark & & 26.7 & 34.1 & 39.7 & 33.5 & - & 39.4 \\
SparseOcc (8f) \cite{liu2023fully}  &  & \checkmark & & 29.1 & 35.8 & 40.3 & 35.1 & - & 30.6 \\
BEVDet-Occ-flow (8f)   & \checkmark & \checkmark & \checkmark & 26.6 & 32.9 & 38.6 & 32.5 & 0.545 & 39.3 \\
FB-Occ-flow (16f)    & \checkmark & \checkmark & \checkmark & 27.3 & 34.3 & 38.9 & 33.5 & 0.505  & 39.2 \\
\midrule
FB-Occ-flow + Ours    & \checkmark & \checkmark & \checkmark & 30.2 & 37.8 & 42.7 & 36.9 \improveb{3.4} & 0.303 \improveb{0.202} & 42.3 \improveb{3.1} \\
\bottomrule
\end{tabular}
\caption{The 3D occupancy prediction performance on the nuScenes validation set is evaluated. The RayIoU and mAVE results are obtained using the annotations from OpenOcc \cite{liu2023fully, sima2023_occnet}, while the mIoU results are based on the Occ3D annotations \cite{tian2024occ3d}. BEVDet-Occ-flow and FB-Occ-flow represent the models with our scene flow decoder integrated into the original architectures.
}
\label{tab:performance}
\end{table*}

\begin{table*}[ht] %
    \setlength{\tabcolsep}{0.003\linewidth}   
    \renewcommand\arraystretch{1.3}
    \centering
    \resizebox{\textwidth}{!}{
    \begin{tabular}{l|c c | c c c c c c c c c c c c c c c c}
        \toprule
        Method
        &  \makecell{SC\\ IoU} & \makecell{SSC \\ mIoU}
        & \rotatebox{90}{ barrier}
        & \rotatebox{90}{ bicycle}
        & \rotatebox{90}{ bus}
        & \rotatebox{90}{ car}
        & \rotatebox{90}{ const. veh.}
        & \rotatebox{90}{ motorcycle}
        & \rotatebox{90}{ pedestrian}
        & \rotatebox{90}{ traffic cone}
        & \rotatebox{90}{ trailer}
        & \rotatebox{90}{ truck}
        & \rotatebox{90}{ drive. suf.}
        & \rotatebox{90}{ other flat}
        & \rotatebox{90}{ sidewalk}
        & \rotatebox{90}{ terrain}
        & \rotatebox{90}{ manmade}
        & \rotatebox{90}{ vegetation}
        \\
        \midrule 
        
        BEVFormer~\cite{li2022bevformer} & 30.50 & 16.75 & 14.22 &	6.58 & 23.46 & 28.28& 8.66 &10.77& 6.64& 4.05& 11.20&	17.78 & 37.28 & 18.00 & 22.88 & 22.17 & 13.80 &	22.21\\
        
        TPVFormer~\cite{huang2023tri} & 11.51 & 11.66 & 16.14&	7.17& 22.63	& 17.13 & 8.83 & 11.39 & 10.46 & 8.23&	9.43 & 17.02 & 8.07 & 13.64 & 13.85 & 10.34 & 4.90 & 7.37\\

        OccFormer~\cite{zhang2023occformer} & 31.39 & 19.03 & 18.65 & 10.41 & 23.92 & 30.29 & 10.31 & 14.19 & 13.59 & 10.13 & 12.49 & 20.77 & 38.78 & 19.79 & 24.19 & 22.21 & 13.48 & 21.35\\
        
        SurroundOcc~\cite{wei2023surroundocc} & 31.49 & 20.30  & 20.59 & 11.68 & 28.06 & 30.86 & 10.70 & 15.14 & 14.09 & 12.06 & 14.38 & 22.26 & 37.29 & 23.70 & 24.49 & 22.77 & 14.89 & 21.86  \\ 

        GaussianFormer~\cite{huang2024gaussian} & 29.83 & 19.10 & 19.52 & 11.26 & 26.11 & 29.78 & 10.47 & 13.83 & 12.58 & 8.67 & 12.74 & 21.57 & 39.63 & 23.28 & 24.46 & 22.99 & 9.59 & 19.12 \\

        FB-OCC~\cite{li2023fb} & 32.37 & 18.68 & 18.22 & 11.04 & 24.30 & 28.63 & 8.86 & 11.27 & 13.66 & 8.71 & 7.99 & 20.35 & 40.45 & 20.86 & 25.73 & 23.59 & 12.67 & 22.48 \\

        \midrule

        FB-OCC + Ours & 35.85 & 22.64 & 22.06 & 14.27 & 27.13 & 31.29 & 14.43 & 17.10 & 15.61 & 12.90 & 14.72 & 24.37 & 44.06 & 26.58 & 28.55 & 27.03 & 16.05 & 26.08 \\
        
        \bottomrule
    \end{tabular}}
    \caption{3D semantic occupancy prediction results on nuScenes validation set with the annotation of OpenOccuancy \cite{wang2023openoccupancy}.}
    \label{tab:nuscseg}
    
\end{table*}

\begin{figure*}[h]
    \centering
    \begin{overpic}[width=\linewidth]{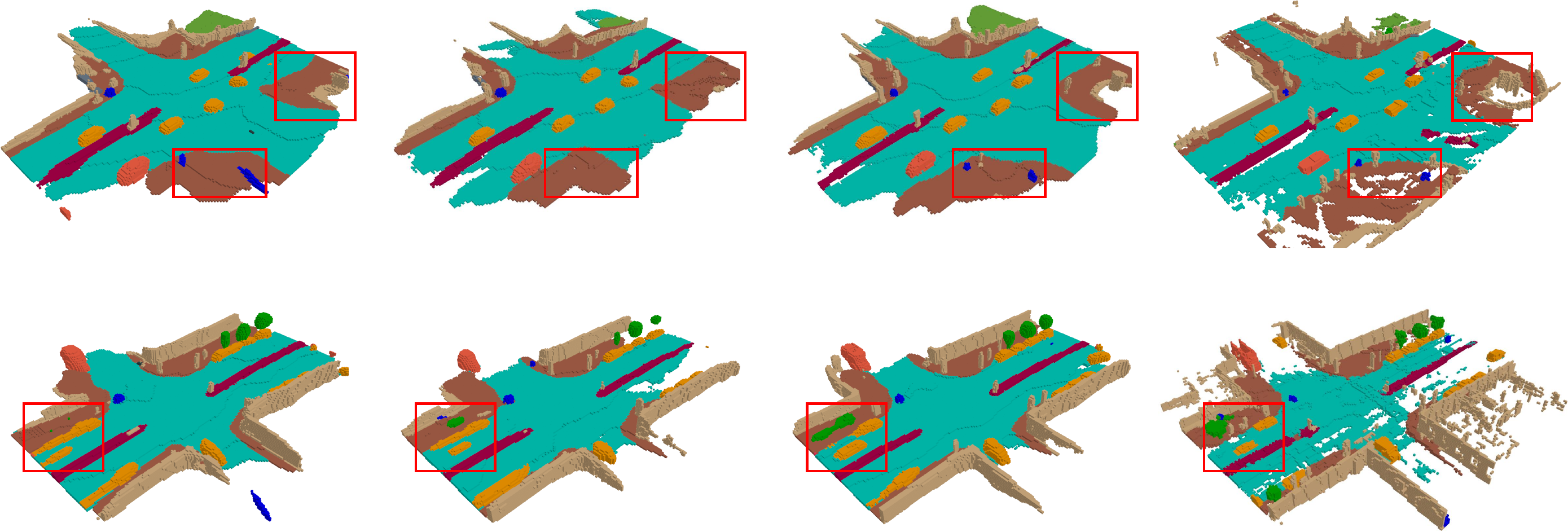}
    \put(5, -2){\small{BEVDet-Occ \cite{huang2021bevdet}}}
    \put(33, -2){\small{FB-Occ \cite{li2023fb}}}
    \put(60, -2){\small{Ours}}
    \put(82, -2){\small{Ground Truth}}
    \end{overpic}\vspace{7pt}
    \caption{The qualitative comparison of our occupancy prediction with other methods is presented. 
    We highlight the regions where our method shows clear superiority using red boxes, emphasizing the areas where the performance differences are most noticeable.}
    \label{fig:compare}
\end{figure*}

\noindent\textbf{Qualitative Comparison.}
In \figref{fig:compare}, 
we show the qualitative results of our method alongside BEVDet-Occ \cite{huang2021bevdet} and FB-Occ \cite{li2023fb}. 
In the first row, 
we observe that our predictions closely match the ground truth in complex street scenes. Comparing the ground truth with the different prediction results in the upper red boxes, 
we can see that our predicted street structure is more accurate than both BEVDet-Occ and FB-Occ. 
In the middle boxes, 
our method successfully predicts the two small blue regions representing pedestrians, 
while FB-Occ fails to identify the pedestrians, and BEVDet-Occ mispredicts their shape. 
In the lower boxes of the second row, two cars are marked by yellow regions. 
Our method accurately predicts both the location and shape of the cars, 
while the other methods fail to predict the car length correctly.

\begin{figure*}[h]
    \centering
    \begin{overpic}[width=\linewidth]{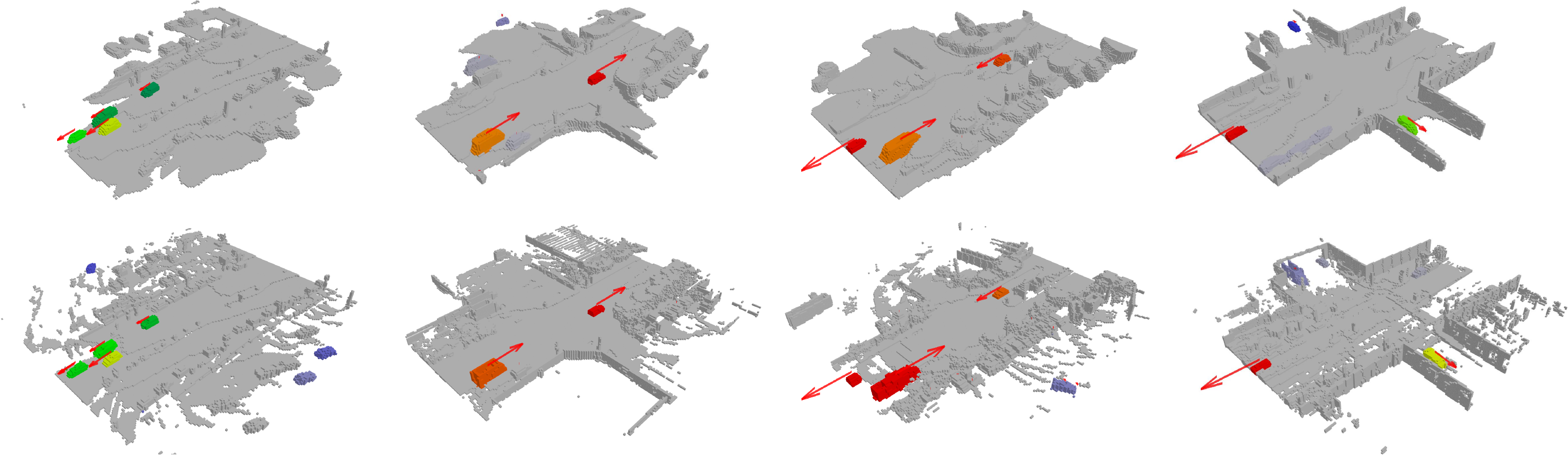}
    \put(-0.4, 3){\rotatebox{90}{\small{Ground Truth}}}
    \put(-0.4, 20){\rotatebox{90}{\small{Ours}}}
    \end{overpic}
    \caption{We present a qualitative comparison of our occupancy prediction with the ground truth. We use a color scale to represent the magnitude of the flow. 
    Red arrows are employed to indicate both the direction and magnitude of the flow.}
    \label{fig:flow}
\end{figure*}

In \figref{fig:flow}, 
we visualize the predicted and ground truth scene flow. In the leftmost scene, four cars are driving closely together, 
and our method accurately predicts the size and direction of the scene flows for each vehicle. In the third and fourth scenes, 
the vehicles are moving in different directions and are far from the ego vehicle. Despite this, our method still accurately predicts their forward directions, 
with only a small deviation in speed prediction. 
These visual results show that our method performs effectively in scene flow prediction and is suitable for real-world applications.
More results are available in the supplementary.

\subsection{Ablation Study}
In this section, we validate the effectiveness of all proposed designs. Specifically, we evaluate: (1) the performance improvement of our method on different models. (2) the superiority of 3D Gaussians over traditional volume rendering \cite{pan2023renderocc}, (3) the impact of dynamic and static decomposition on the learning of scene flow, and (4) how weight point sampling addresses the issue of class imbalance. 

\begin{table}[ht]
\centering
\small
\resizebox{0.45\textwidth}{!}{\begin{tabular}{l|ccc|c|c}
\toprule
Methods & \multicolumn{3}{c|}{RayIoU$_{1, 2, 4m}$ $\uparrow$}  & RayIoU $\uparrow$ & mAVE $\downarrow$  \\ 
\midrule
BEVDet-Occ \cite{huang2021bevdet}  & 26.0 & 32.4 & 38.2 & 32.0 & -  \\
+ Ours     & 29.2 & 36.4 & 41.1 & 35.6 \improveb{3.6}  & 0.314  \\
\midrule
FB-Occ \cite{li2023fb}     & 26.7 & 34.1 & 39.7 & 33.5 & -  \\
+ Ours     & 30.2 & 37.8 & 42.7 & 36.9 \improveb{3.4} & 0.303  \\
\midrule
SparseOcc  \cite{liu2023fully}  & 29.1 & 35.8 & 40.3 & 35.1 & -  \\
+ Ours     & 31.5 & 38.9 & 43.5 & 38.0 \improveb{2.9} & 0.301  \\


\bottomrule
\end{tabular}}
\caption{Improvement on different frameworks.}
\label{tab:improve}
\end{table}

\noindent
\textbf{Improvement on different models.}
In \tabref{tab:improve}, 
we show the efficacy and generality of the proposed VoxelSplat framework on three popular occupancy models: BEVDet-Occ \cite{huang2021bevdet}, FB-Occ \cite{li2023fb} and SparseOcc~\cite{liu2023fully}.

By integrating our VoxelSplat into advanced models, the occupancy prediction performance of BEVDet-Occ \cite{huang2021bevdet}, FB-Occ \cite{li2023fb}, and SparseOcc~\cite{liu2023fully} improves by 3.6, 3.4, and 2.9, 
respectively. The scene flows are also accurately predicted, with mAVE values of 0.314, 0.303, and 0.301.

\begin{table*}[ht]     
\centering
\small
\begin{tabular}{c|ccccc|ccc|c|c}
\toprule
 & Nerf & Gaussian & Multi-frames  & Decompose & WS & \multicolumn{3}{c|}{RayIoU$_{1m, 2m, 4m}$ $\uparrow$} & RayIoU $\uparrow$ & mAVE $\downarrow$ \\ 
\midrule
Version & \multicolumn{5}{c|}{BEVDet-Occ-flow (8f)}   & 26.6 & 32.9 & 38.6 & 32.5 & .545 \\
 \midrule
A & \checkmark &  &  &  &  & 26.9 & 33.7 & 38.2 & 32.9 \improveb{0.4} & .569 \decrease{.024}  \\
B &  & \checkmark &  &  &  & 27.5 & 34.5 & 39.1 & 33.6 \improveb{1.1} & .515 \improveb{.030}  \\
C &  & \checkmark & \checkmark &  &  & 27.9 & 35.2 & 39.3 & 34.1 \improveb{1.6}  & .487 \improveb{.068}  \\
D &  &  & \checkmark & \checkmark &  & 25.3 & 32.1 & 36.8 & 31.4 \decrease{1.1} & .792 \decrease{.247}  \\
E &  & \checkmark & \checkmark & \checkmark & & 27.6  & 35.5 & 39.8 & 34.3 \improveb{1.8} & .353 \improveb{.192} \\
F &  & \checkmark & \checkmark & \checkmark & \checkmark & 29.2 & 36.4 & 41.1 & 35.6 \improveb{3.1} & .314 \improveb{.231} \\
\bottomrule
\end{tabular}
\caption{Ablation Study our method. Nerf denotes the supervision of volume rendering \cite{mildenhall2021nerf}.
Multi-frames denotes using the 2D GT of adjust frames. Decompose denotes the separate supervison of dynamic and static objects.
WS denotes our Weighted Sampling strategy.}
\label{tab:ablation}
\vspace{-15pt}
\end{table*}
\noindent
\textbf{Comparison with Volume Rendering.}
The strategy of Volume Rendering \cite{mildenhall2021nerf, pan2023renderocc} can also predict 2D camera view semantics and depths for auxiliary supervision. 
In versions A and B of \tabref{tab:ablation}, we compare volume rendering with our Semantics Gaussian Splatting.
The results show that volume rendering leads to a performance improvement of 0.4, but a 0.024 decrease in mAVE.
In contrast, our Semantics Gaussian Splatting achieves a significant improvement of 1.1 in RayIoU and 0.03 in mAVE.

Unlike volume rendering, which densely samples points along many camera rays, our Gaussians are primarily decoded from occupied voxels.
Since most sampled points are located in empty voxels, volume rendering may focus too much on empty space. 
In contrast, our Semantics Gaussians concentrate on learning from the occupied space.
As a result, the 3D Gaussian Splatting strategy provides more benefits for the occupancy prediction task.

\noindent
\textbf{Effects of Decomposition.} 
In version C of \tabref{tab:ablation}, we add virtual camera views from future time stamps to boost training. 
This leads to an improvement in RayIoU from 33.6 to 34.1, and a reduction in mAVE from 0.515 to 0.487.

In version E of \tabref{tab:ablation}, our VoxelSplat separately supervises the rendering results of static and two-frame dynamic objects. 
The results show a significant improvement in scene flow prediction, with mAVE improving from 0.487 to 0.353. This demonstrates that this strategy helps the model focus more on learning of dynamic objects.

\begin{figure}[ht]
    \centering
    \begin{subfigure}[b]{0.46\textwidth}
        \centering
        \includegraphics[width=\linewidth]{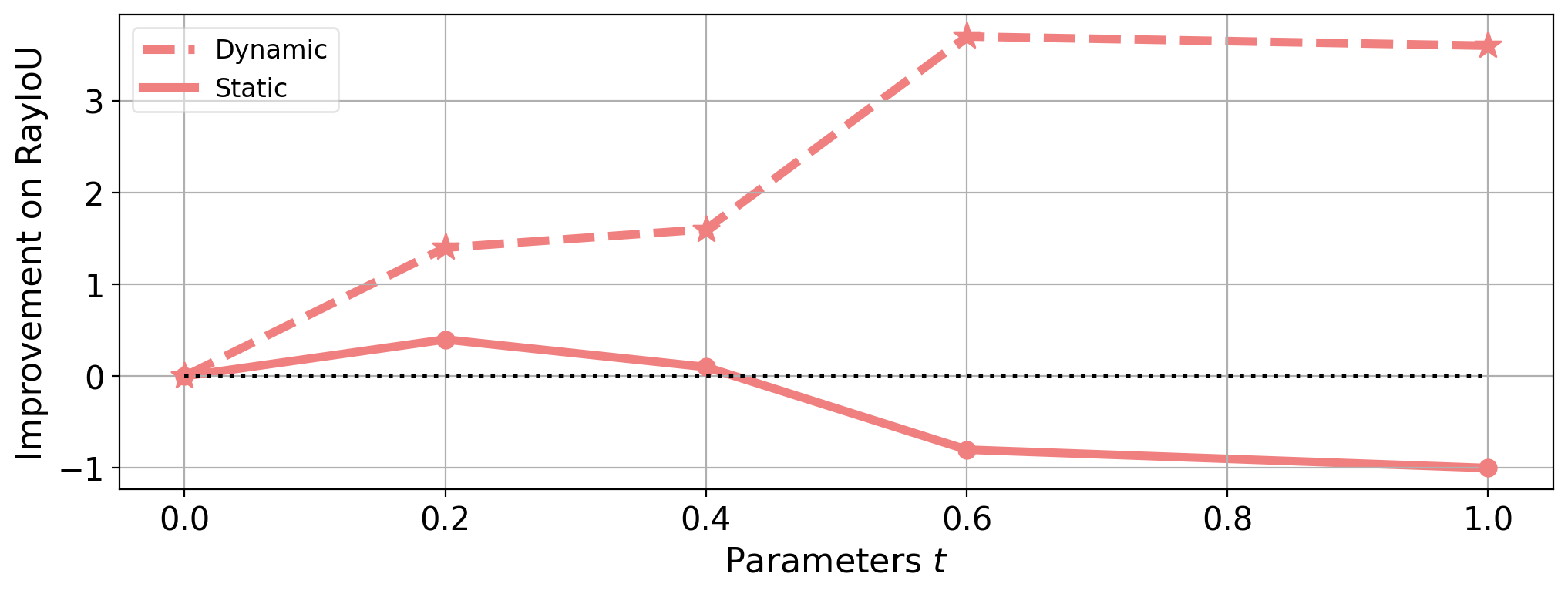} 
        \label{fig:sub1}
    \end{subfigure}
    \hspace{0pt} 
    \begin{subfigure}[b]{0.46\textwidth}
        \centering
        \includegraphics[width=\textwidth]{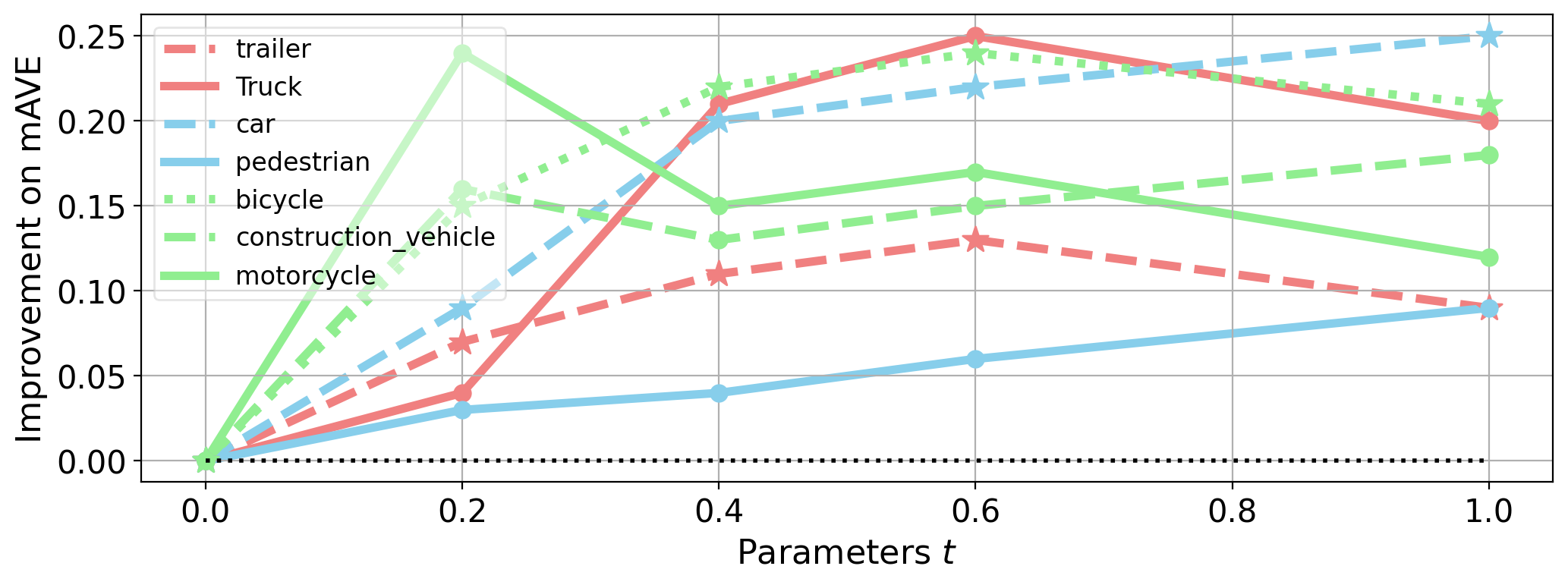} 
        \label{fig:sub2}
     
    \end{subfigure}
    \caption{The effect of sampling function hyperparameter $t$ on perception performance.}
    \label{fig:t}
    \vspace{-18pt}
\end{figure}

\begin{figure}[ht]
    \centering
    \begin{subfigure}[b]{0.46\textwidth}
        \centering
        \begin{overpic}[width=\textwidth]{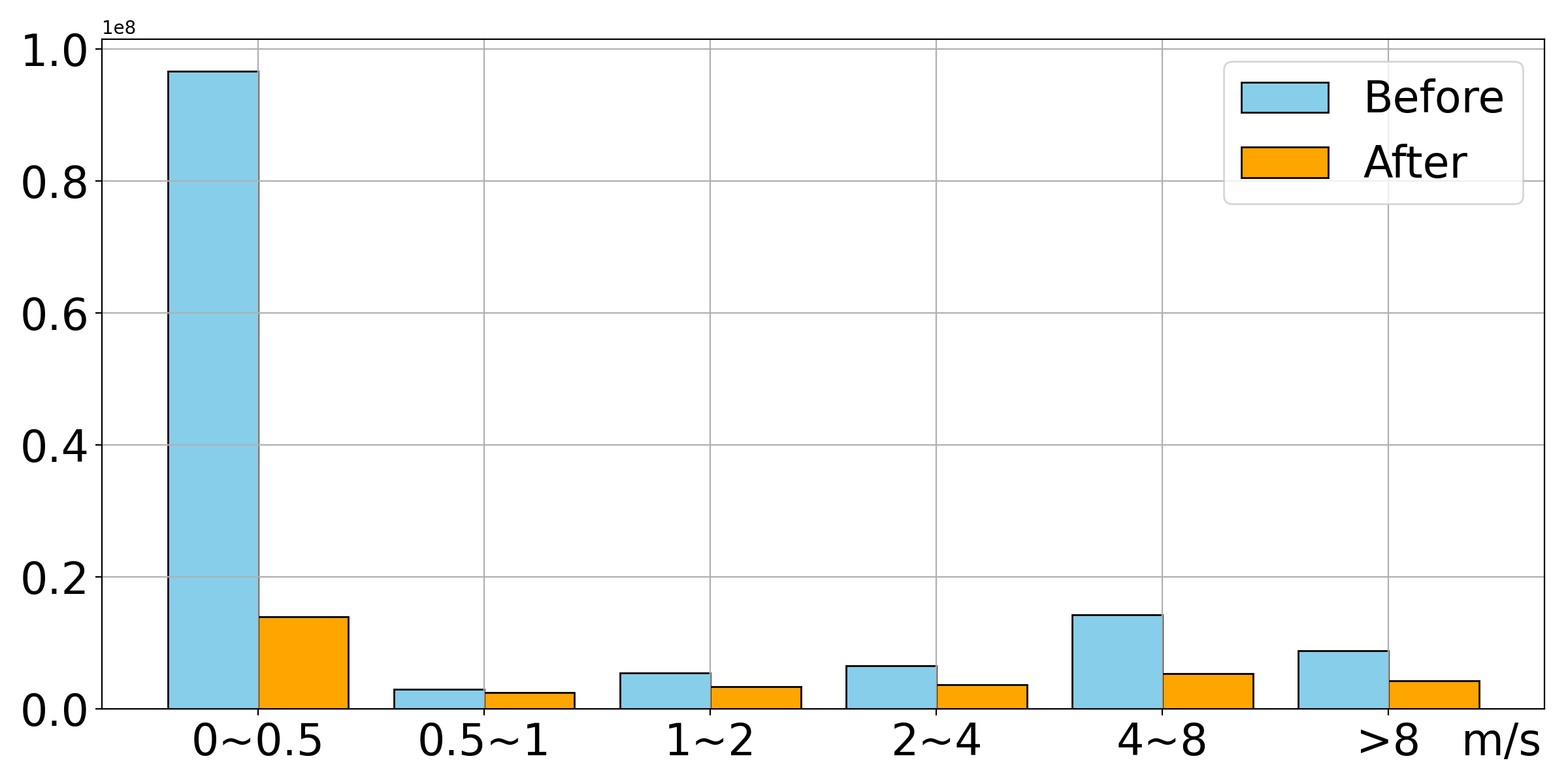}
            \put(26, -2){\small{Distribution of scene flow magnitude}}  
        \end{overpic}
        \vspace{5pt} 
    \end{subfigure}
    \begin{subfigure}[b]{0.46\textwidth}
        \centering
        \begin{overpic}[width=\textwidth]{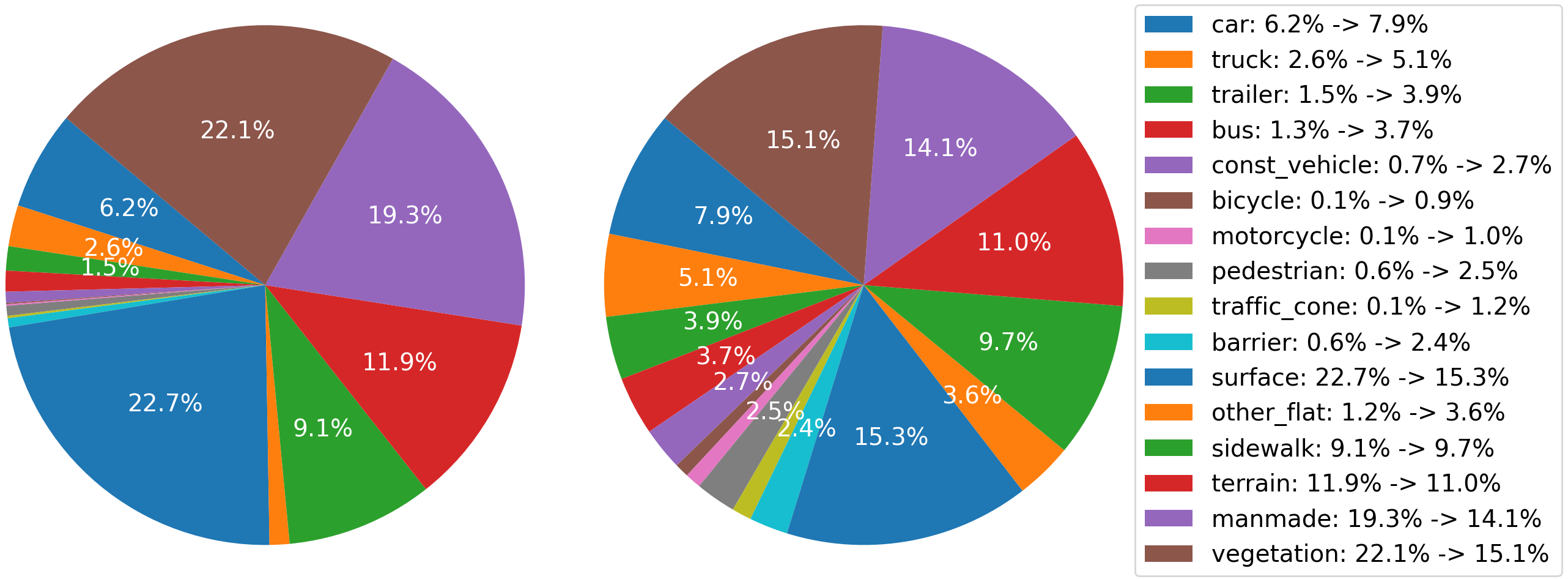}
            \put(35, -4){\small{Distribution of semanics}}  
        \end{overpic}
    \end{subfigure}
    \caption{The semantic and scene flow magnitude distributions of OpenOcc \cite{liu2023fully, sima2023_occnet} are shown. 
    The histogram depicts voxel velocity distribution in the dynamic object class before and after weighted sampling. 
    The pie chart shows category distributions before and after sampling.}
    \label{fig:ratio}
    \vspace{-18pt}
\end{figure}

\noindent
\textbf{Effects of Weight Points Sampling.} 
The weighted point sampling strategy is designed to address the issues of class imbalance and large speed variations in occupancy and flow prediction. As shown in \figref{fig:ratio}, in OpenOcc \cite{caesar2020nuscenes, sima2023_occnet}, the speeds of most voxels corresponding to dynamic objects are lower than 0.5 m/s, making it difficult for the model to capture the motion of these objects. At the same time, most voxels belong to static objects (\eg vegetation, manmade structures, and sidewalks). The ratios of voxels corresponding to pedestrians, bicycles, and motorcycles are even lower than 1\%, which causes the model to pay less attention to these important classes, despite their significant roles in driving decisions.

To address these challenges, we employ the weighted point sampling function defined in \eqnref{eq:probability}, controlled by the hyperparameter \( t \). As shown in \figref{fig:t}, we illustrate the influence of \( t \) on performance. When \( t = 0 \), all voxels have equal probabilities of being sampled. As \( t \) increases, the sampling probability of voxels belonging to dynamic objects increases, leading to improved performance in both RayIoU and mAVE. However, when \( t \) exceeds 0.5, the performance on dynamic classes no longer improves, while the performance on static classes starts to degrade. Based on this analysis, we set \( t = 0.5 \) in Version F of \tabref{tab:ablation}, resulting in performance improvements of 3.1 and 0.231 in RayIoU and mAVE, respectively.

Additionally, we explore to apply the decomposition and weighted sampling strategy directly to the 3D loss, without incorporating the 2D loss. 
As shown in Version E of \tabref{tab:ablation}, this approach leads to a significant performance drop. We hypothesize that this is due to the lack of supervisory signals in certain areas of the scene.



\section{Conclusion and Limitation}

In this work, we propose VoxelSplat, a novel Semantic Gaussian Splatting framework, to explore the potential of 4D Gaussians for occupancy and flow prediction. Our focus is on the Gaussian rendering loss, with Dynamic \& Static Decomposition and Weighted Point Sampling designs, which enhance the model's ability to learn occupancy and scene flow. VoxelSplat is a plug-and-play solution that improves the performance of existing occupancy models without increasing inference time. 

However, there is still potential for further research in applying Gaussians to occupancy and flow prediction. We highlight two limitations: (1) Despite the 2D rendering loss aiding self-supervised scene flow learning, our model still requires ground truth 3D scene flow. (2) This work focuses on occupancy and flow prediction but could be extended to other autonomous  tasks,
such as occupancy forecasting.

\clearpage
\setcounter{page}{1}
\maketitlesupplementary

\section{Supplementary}
In the supplementary material, we provide additional details to complement the main paper. These include:

\begin{itemize}
    \item \textbf{Deeper Analysis of Rendering Losses:} An exploration of the impact of rendering losses on the convergence of 3D occupancy and scene flow.
    \item \textbf{Visualization of Rendering Results:} Examples of rendering outputs on the validation set, illustrating what the rendering branch learns after training.
    \item \textbf{Additional Qualitative Results:} A demonstration of the predicted 3D occupancy and scene flow through multiview video visualizations, showcasing the quality of our method.
\end{itemize}

\subsection{Deeper Analysis of Rendering Losses}

\begin{figure}[ht!]
    \centering
    \begin{subfigure}[b]{0.47\textwidth}
        \centering
        \begin{overpic}[width=\textwidth]{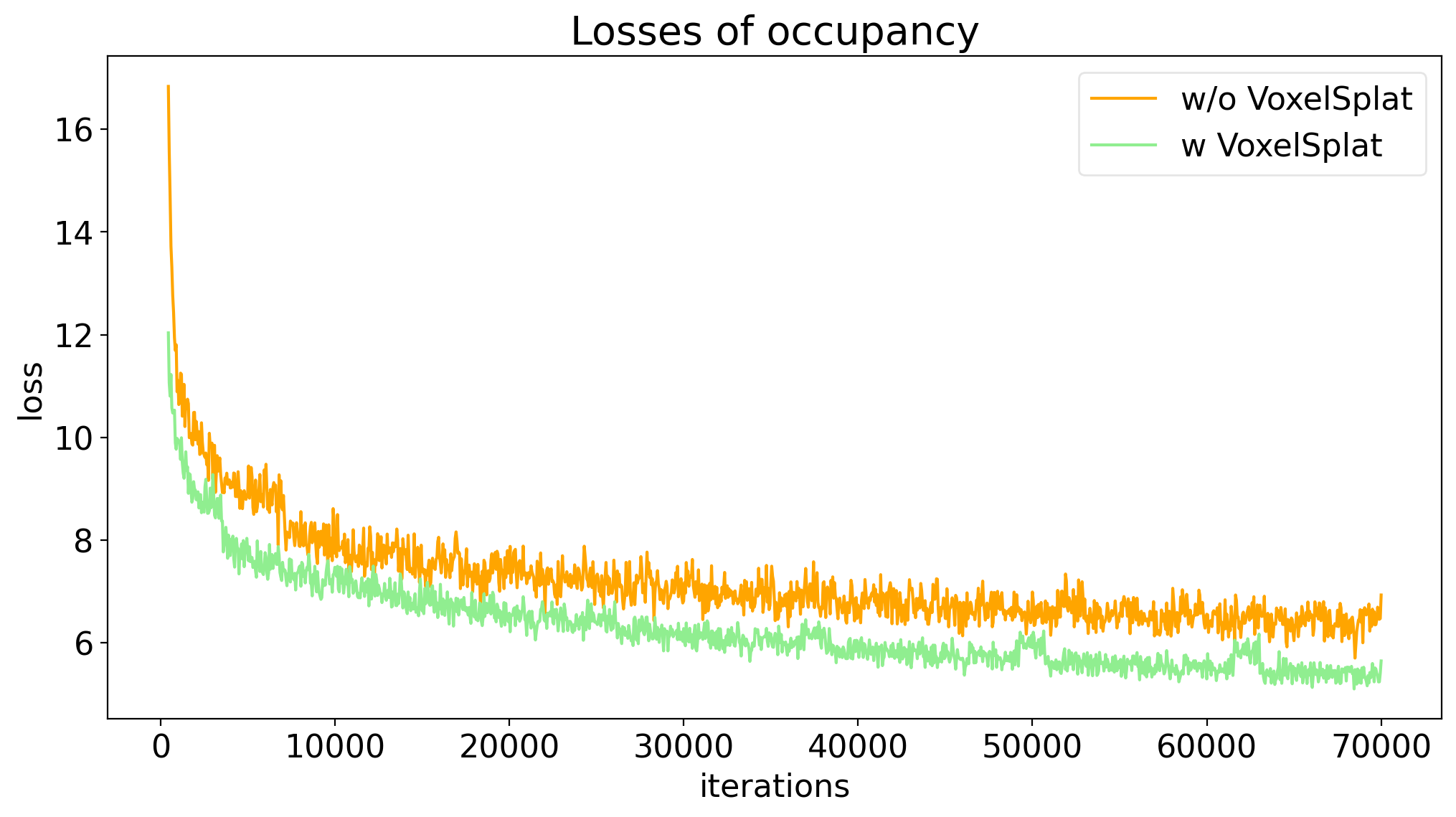}
        \end{overpic}
        \vspace{-5pt}
    \end{subfigure}
    
    \begin{subfigure}[b]{0.47\textwidth}
        \centering
        \begin{overpic}[width=\textwidth]{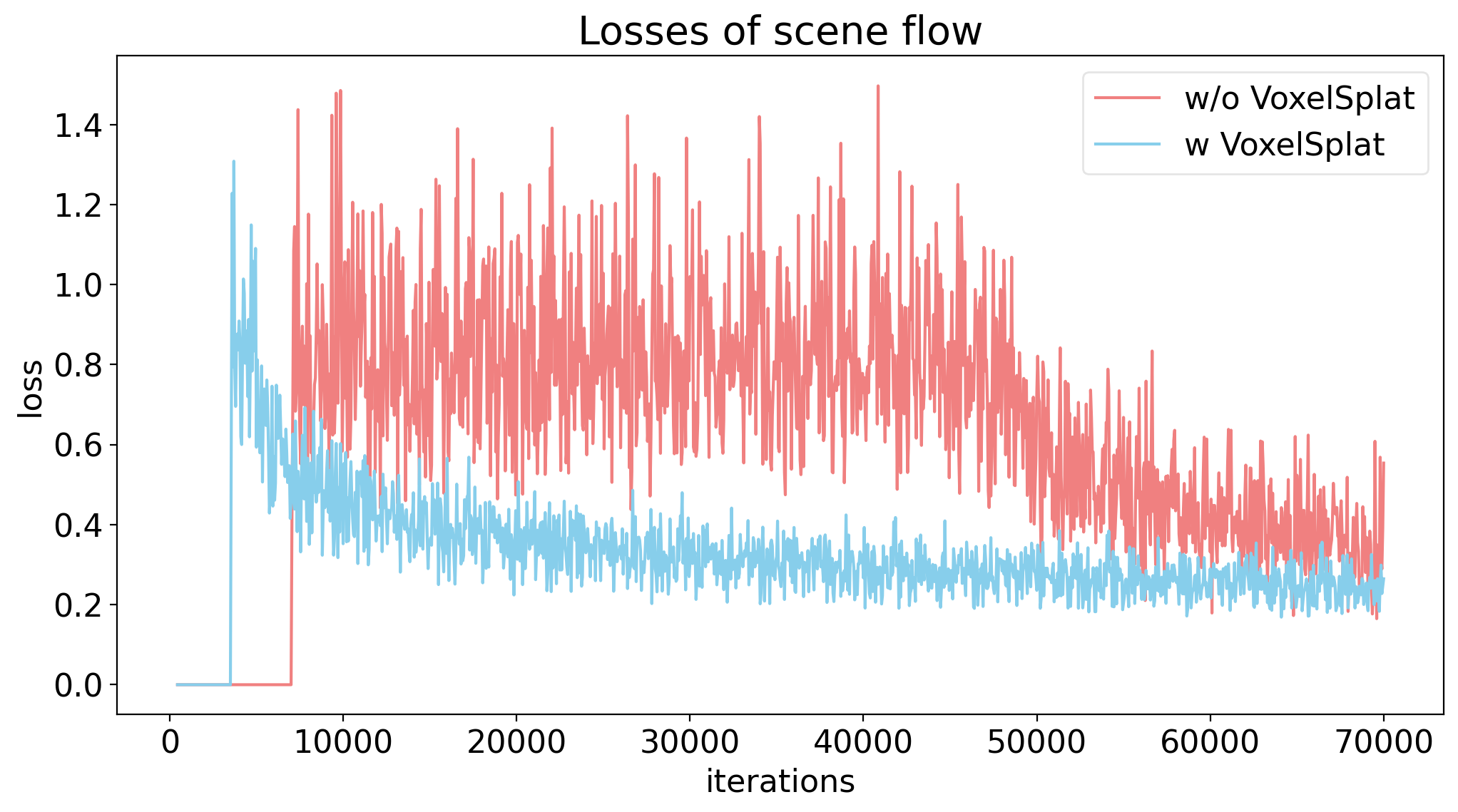}
        \end{overpic}
        \vspace{-5pt}
    \end{subfigure}
    \begin{subfigure}[b]{0.47\textwidth}
        \centering
        \begin{overpic}[width=\textwidth]{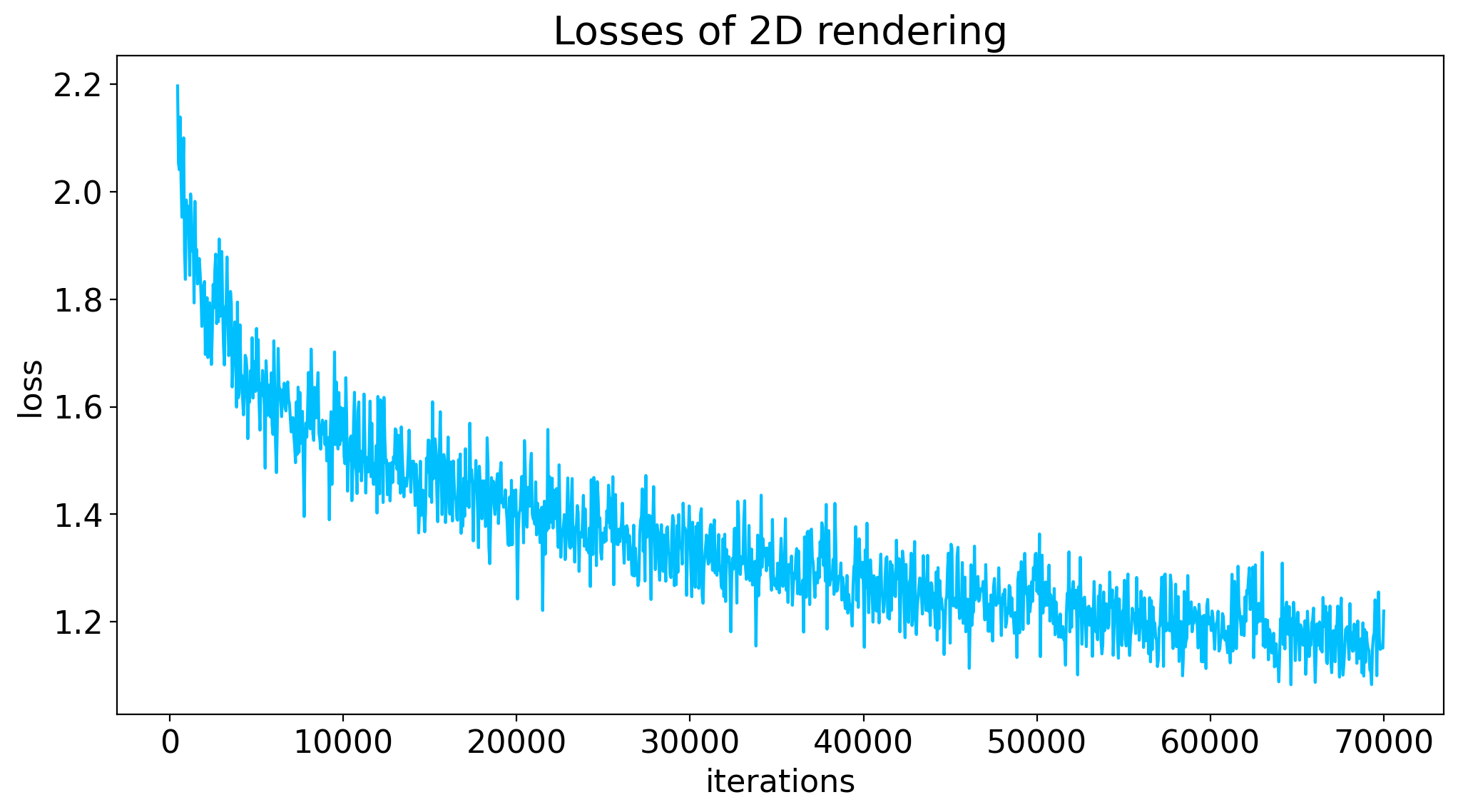}
        \end{overpic}
        \vspace{-5pt}
    \end{subfigure}
    \caption{The comparison of loss curves with and without our VoxelSpat. }
    \label{fig:curve}
\end{figure}

In \figref{fig:curve}, we compare the occupancy and flow loss curves with and without the rendering loss $\mathcal{L}_{2D}$.

\noindent\textbf{Detailed Experimental Settings.}  
We conduct our loss curve experiments based on the model architecture of FB-Occ \cite{li2023fb}. Following the original settings, the occupancy loss $\mathcal{L}_{occ}$ consists of \textit{cross-entropy loss}, \textit{Lovász-Softmax loss} \cite{berman2018lovasz}, and \textit{scaling loss}.  
As mentioned in the main paper, we employ the L1 loss as the scene flow loss function $\mathcal{L}_{flow}$.  
To prevent training collapse, we start computing the flow loss at 3500 iterations, after which FB-Occ begins using temporal information.  
We train the model with and without our rendering loss $\mathcal{L}_{2D}$ for 70,000 iterations and compare the convergence of the loss curves.

\noindent\textbf{Effect of Rendering on 3D Losses.}  
From the upper figure in \figref{fig:curve}, we observe that $\mathcal{L}_{occ}$ converges faster with the inclusion of $\mathcal{L}_{2D}$.  
In the middle figure, the flow loss $\mathcal{L}_{flow}$ starts converging after 40,000 iterations. This is likely due to the small proportion of dynamic objects in the scenes, which makes it challenging for the model to capture motion information.  
However, with our $\mathcal{L}_{2D}$, which specifically addresses dynamic objects, the $\mathcal{L}_{flow}$ converges significantly faster.

This experiment demonstrates that our strategy of explicit modeling of the occupancy field with 3D Gaussians and splat rendering supervision helps the original loss functions find a better convergence direction.

\subsection{Visualization of Rendering Results}

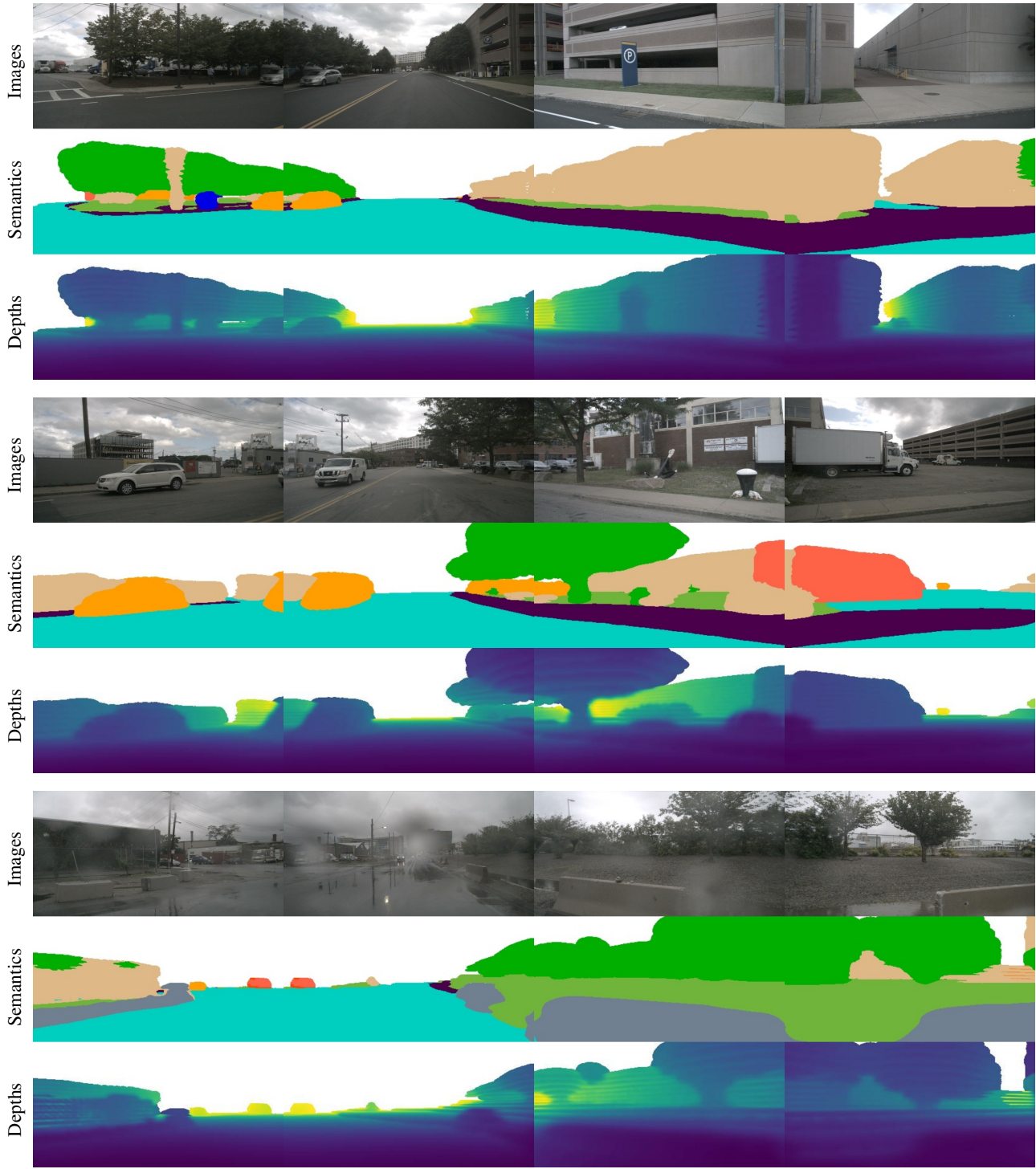
\begin{figure*}[t!]
    \centering
    \begin{subfigure}[b]{0.95\textwidth}
        \centering
        \begin{overpic}[width=\textwidth]{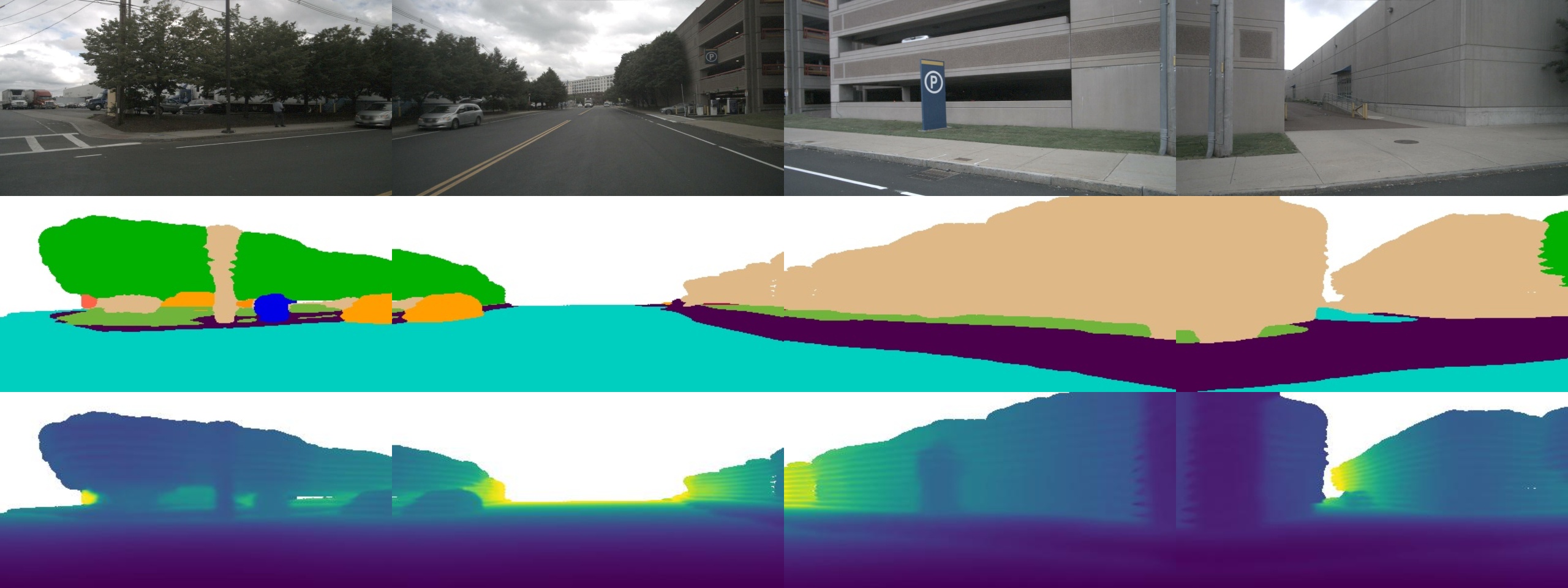}
        \put(-2.5, 28){\rotatebox{90}{\small{Images}}}
        \put(-2.5, 14){\rotatebox{90}{\small{Semantics}}}
        \put(-2.5, 3){\rotatebox{90}{\small{Depths}}}
        \end{overpic}
        \vspace{-2pt}
    \end{subfigure}
    \begin{subfigure}[b]{0.95\textwidth}
        \centering
        \begin{overpic}[width=\textwidth]{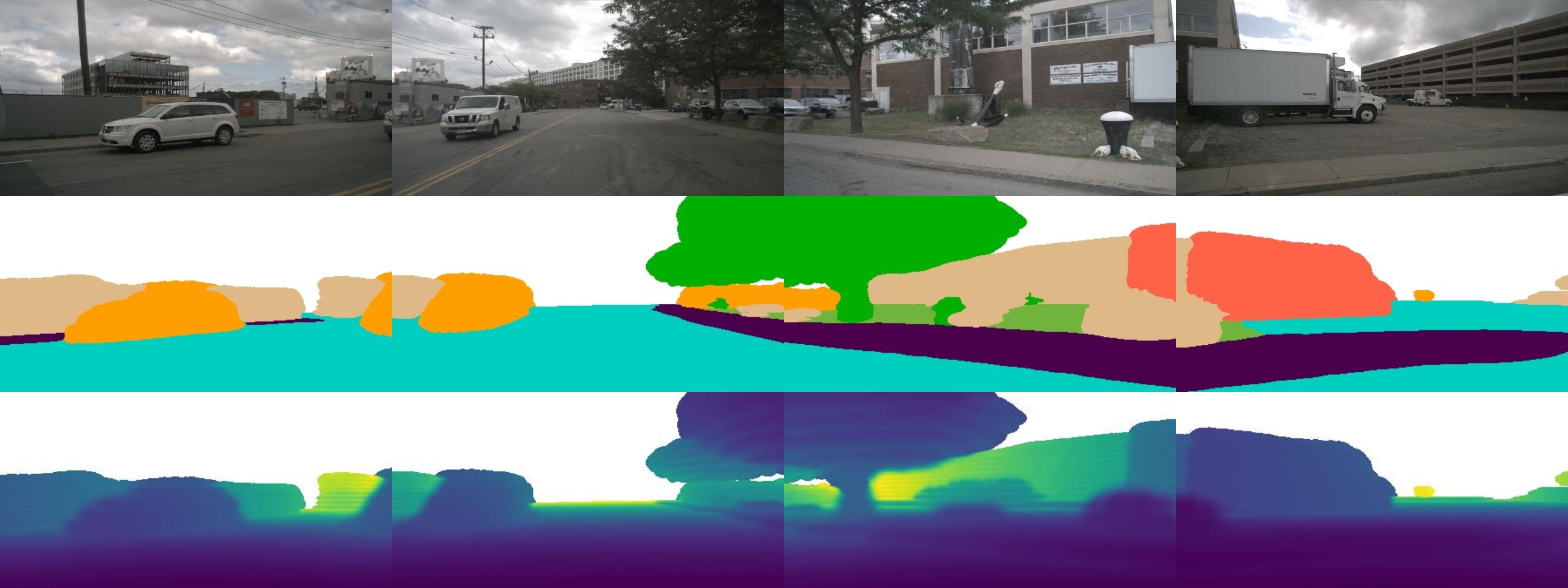}
        \put(-2.5, 28){\rotatebox{90}{\small{Images}}}
        \put(-2.5, 14){\rotatebox{90}{\small{Semantics}}}
        \put(-2.5, 3){\rotatebox{90}{\small{Depths}}}
        \end{overpic}
        \vspace{-2pt}
    \end{subfigure}
    \begin{subfigure}[b]{0.95\textwidth}
        \centering
        \begin{overpic}[width=\textwidth]{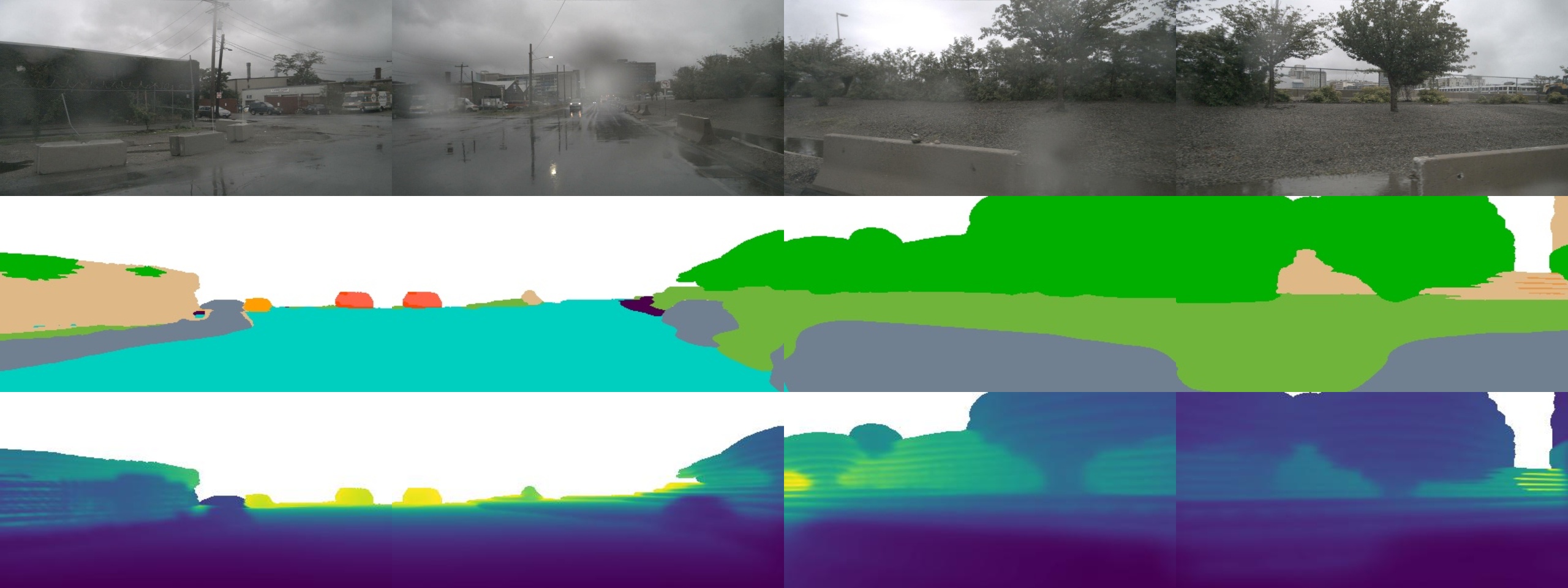}
        \put(-2.5, 28){\rotatebox{90}{\small{Images}}}
        \put(-2.5, 14){\rotatebox{90}{\small{Semantics}}}
        \put(-2.5, 3){\rotatebox{90}{\small{Depths}}}
        \end{overpic}
    \end{subfigure}
    \caption{The visualization results of rendering semantics and depths on the validation dataset \cite{caesar2020nuscenes} are presented.}
    \label{fig:rendering}
\end{figure*}

Although our rendering branch is not used during inference, we conduct a simple visualization experiment on the validation set of \cite{caesar2020nuscenes} to help understand what the rendering branch learns during training.  
Specifically, based on FB-Occ \cite{li2023fb}, 640,000 semantic Gaussians initialized from all voxel centers are predicted by the decoder in the rendering branch.  
Gaussians with opacity higher than 0.2 are splatted into the camera view.  
The rendering semantics and depth results in \figref{fig:rendering} demonstrate that our rendering branch successfully predicts high-quality semantics and depths, even under adverse weather conditions.

\subsection{Additional Qualitative Results}
\begin{figure*}[t!]
    \centering
    \begin{subfigure}[b]{\textwidth}
        \centering
        \begin{overpic}[width=\textwidth]{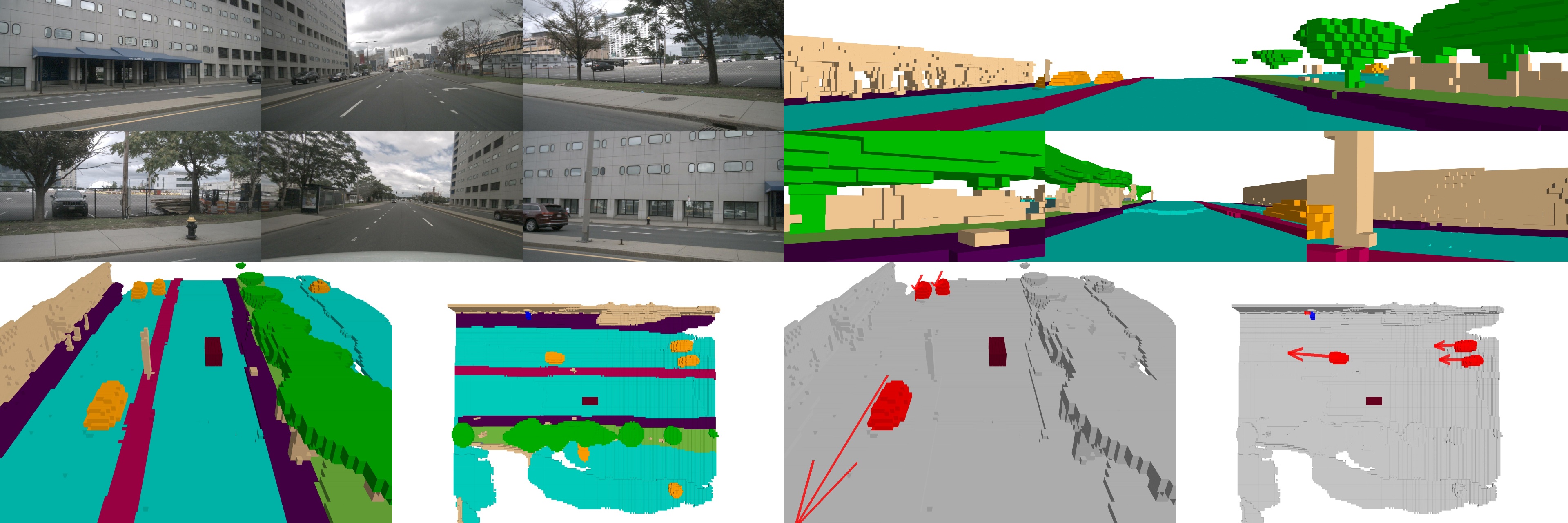}
        \end{overpic}
        \vspace{0pt}
    \end{subfigure}
    
    \begin{subfigure}[b]{\textwidth}
        \centering
        \begin{overpic}[width=\textwidth]{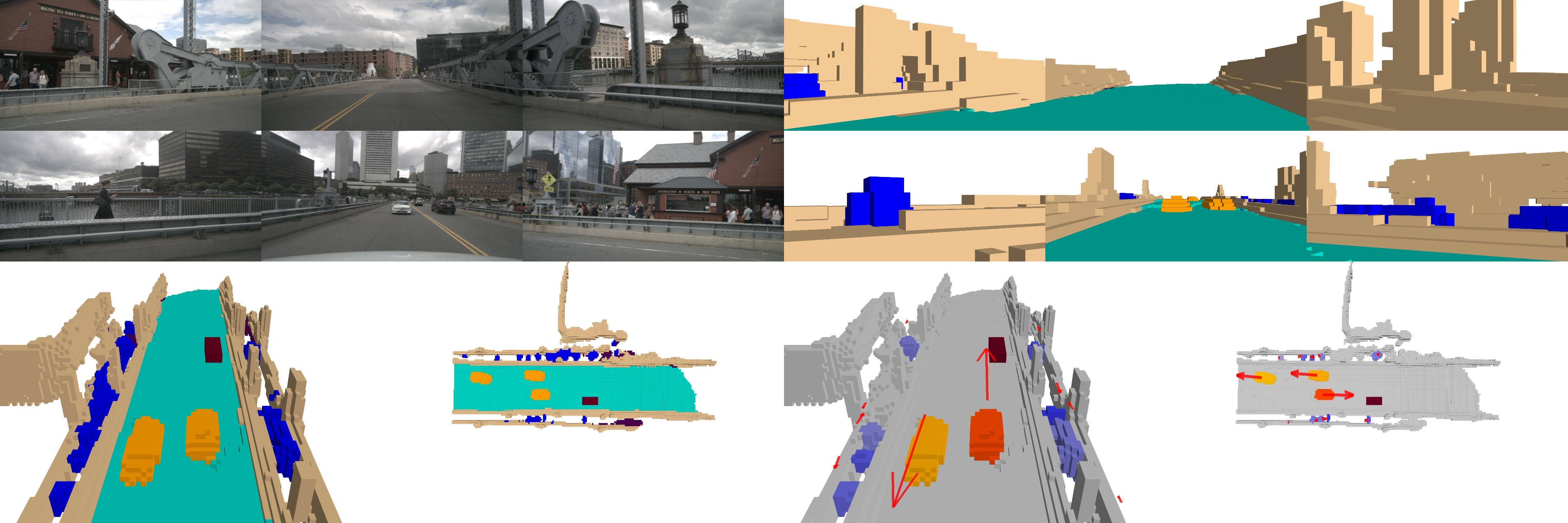}
        \end{overpic}
        \vspace{0pt}
    \end{subfigure}
    \begin{subfigure}[b]{\textwidth}
        \centering
        \begin{overpic}[width=\textwidth]{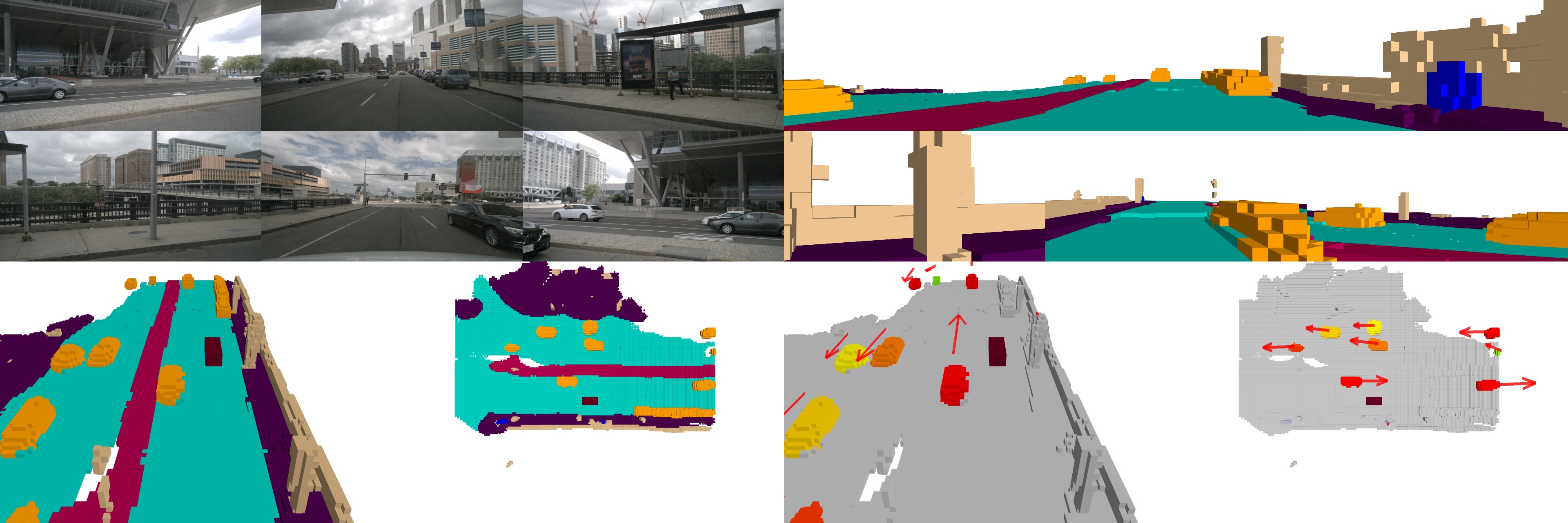}
        \end{overpic}
    \end{subfigure}
    \caption{We provide a series of videos in the supplementary material, which demonstrate the predicted occupancy and flow from different viewpoints.
  }
    \label{fig:video}
\end{figure*}

In \figref{fig:video}, 
we illustrate the image inputs and a more comprehensive visualizations of our predicted occupancy and flow from different viewpoints. 
Further, a series of videos are provided in the supplementary material to validate our accuracy and stability,
which is crucial in self-driving safety.

{
    \small
    \bibliographystyle{ieeenat_fullname}
    \bibliography{main}

\begin{thebibliography}{61}
\providecommand{\natexlab}[1]{#1}
\providecommand{\url}[1]{\texttt{#1}}
\expandafter\ifx\csname urlstyle\endcsname\relax
  \providecommand{\doi}[1]{doi: #1}\else
  \providecommand{\doi}{doi: \begingroup \urlstyle{rm}\Url}\fi

\bibitem[Bae et~al.(2024)Bae, Kim, Yun, Lee, Bang, and Uh]{bae2024per}
Jeongmin Bae, Seoha Kim, Youngsik Yun, Hahyun Lee, Gun Bang, and Youngjung Uh.
\newblock Per-gaussian embedding-based deformation for deformable 3d gaussian splatting.
\newblock \emph{arXiv preprint arXiv:2404.03613}, 2024.

\bibitem[Berman et~al.(2018)Berman, Triki, and Blaschko]{berman2018lovasz}
Maxim Berman, Amal~Rannen Triki, and Matthew~B Blaschko.
\newblock The lov{\'a}sz-softmax loss: A tractable surrogate for the optimization of the intersection-over-union measure in neural networks.
\newblock In \emph{Proceedings of the IEEE conference on computer vision and pattern recognition}, pages 4413--4421, 2018.

\bibitem[Blinn(1982)]{blinn1982generalization}
James~F Blinn.
\newblock A generalization of algebraic surface drawing.
\newblock \emph{ACM transactions on graphics (TOG)}, 1\penalty0 (3):\penalty0 235--256, 1982.

\bibitem[Caesar et~al.(2020)Caesar, Bankiti, Lang, Vora, Liong, Xu, Krishnan, Pan, Baldan, and Beijbom]{caesar2020nuscenes}
Holger Caesar, Varun Bankiti, Alex~H Lang, Sourabh Vora, Venice~Erin Liong, Qiang Xu, Anush Krishnan, Yu Pan, Giancarlo Baldan, and Oscar Beijbom.
\newblock nuscenes: A multimodal dataset for autonomous driving.
\newblock In \emph{Proceedings of the IEEE/CVF conference on computer vision and pattern recognition}, pages 11621--11631, 2020.

\bibitem[Chen et~al.(2024)Chen, Yang, Huang, de~Lutio, Esturo, Ivanovic, Litany, Gojcic, Fidler, Pavone, et~al.]{chen2024omnire}
Ziyu Chen, Jiawei Yang, Jiahui Huang, Riccardo de Lutio, Janick~Martinez Esturo, Boris Ivanovic, Or Litany, Zan Gojcic, Sanja Fidler, Marco Pavone, et~al.
\newblock Omnire: Omni urban scene reconstruction.
\newblock \emph{arXiv preprint arXiv:2408.16760}, 2024.

\bibitem[Chu et~al.(2024)Chu, Ke, and Fragkiadaki]{chu2024dreamscene4d}
Wen-Hsuan Chu, Lei Ke, and Katerina Fragkiadaki.
\newblock Dreamscene4d: Dynamic multi-object scene generation from monocular videos.
\newblock \emph{arXiv preprint arXiv:2405.02280}, 2024.

\bibitem[Duan et~al.(2024)Duan, Wei, Dai, He, Chen, and Chen]{duan20244d}
Yuanxing Duan, Fangyin Wei, Qiyu Dai, Yuhang He, Wenzheng Chen, and Baoquan Chen.
\newblock 4d gaussian splatting: Towards efficient novel view synthesis for dynamic scenes.
\newblock \emph{arXiv preprint arXiv:2402.03307}, 2024.

\bibitem[Feng et~al.(2024)Feng, Cao, Chen, Mu, Martin, and Hu]{feng2024new}
Qiyuan Feng, Gengchen Cao, Haoxiang Chen, Tai-Jiang Mu, Ralph~R Martin, and Shi-Min Hu.
\newblock A new split algorithm for 3d gaussian splatting.
\newblock \emph{arXiv preprint arXiv:2403.09143}, 2024.

\bibitem[Gan et~al.(2024)Gan, Liu, Xu, Mo, and Yokoya]{gan2024gaussianocc}
Wanshui Gan, Fang Liu, Hongbin Xu, Ningkai Mo, and Naoto Yokoya.
\newblock Gaussianocc: Fully self-supervised and efficient 3d occupancy estimation with gaussian splatting.
\newblock \emph{arXiv preprint arXiv:2408.11447}, 2024.

\bibitem[Gao et~al.(2024)Gao, Xu, Cao, Mildenhall, Ma, Chen, Tang, and Neumann]{gao2024gaussianflow}
Quankai Gao, Qiangeng Xu, Zhe Cao, Ben Mildenhall, Wenchao Ma, Le Chen, Danhang Tang, and Ulrich Neumann.
\newblock Gaussianflow: Splatting gaussian dynamics for 4d content creation.
\newblock \emph{arXiv preprint arXiv:2403.12365}, 2024.

\bibitem[Guo et~al.(2024)Guo, Zhou, Li, Wang, and Li]{guo2024motion}
Zhiyang Guo, Wengang Zhou, Li Li, Min Wang, and Houqiang Li.
\newblock Motion-aware 3d gaussian splatting for efficient dynamic scene reconstruction.
\newblock \emph{arXiv preprint arXiv:2403.11447}, 2024.

\bibitem[Huang et~al.(2021)Huang, Huang, Zhu, Ye, and Du]{huang2021bevdet}
Junjie Huang, Guan Huang, Zheng Zhu, Yun Ye, and Dalong Du.
\newblock Bevdet: High-performance multi-camera 3d object detection in bird-eye-view.
\newblock \emph{arXiv preprint arXiv:2112.11790}, 2021.

\bibitem[Huang et~al.(2023{\natexlab{a}})Huang, Zheng, Zhang, Zhou, and Lu]{huang2023selfocc}
Yuanhui Huang, Wenzhao Zheng, Borui Zhang, Jie Zhou, and Jiwen Lu.
\newblock Selfocc: Self-supervised vision-based 3d occupancy prediction.
\newblock \emph{arXiv preprint arXiv:2311.12754}, 2023{\natexlab{a}}.

\bibitem[Huang et~al.(2023{\natexlab{b}})Huang, Zheng, Zhang, Zhou, and Lu]{huang2023tri}
Yuanhui Huang, Wenzhao Zheng, Yunpeng Zhang, Jie Zhou, and Jiwen Lu.
\newblock Tri-perspective view for vision-based 3d semantic occupancy prediction.
\newblock In \emph{CVPR}, pages 9223--9232, 2023{\natexlab{b}}.

\bibitem[Huang et~al.(2024{\natexlab{a}})Huang, Zheng, Zhang, Zhou, and Lu]{huang2024gaussian}
Yuanhui Huang, Wenzhao Zheng, Yunpeng Zhang, Jie Zhou, and Jiwen Lu.
\newblock Gaussianformer: Scene as gaussians for vision-based 3d semantic occupancy prediction.
\newblock \emph{arXiv preprint arXiv:2405.17429}, 2024{\natexlab{a}}.

\bibitem[Huang et~al.(2024{\natexlab{b}})Huang, Sun, Yang, Lyu, Cao, and Qi]{huang2024sc}
Yi-Hua Huang, Yang-Tian Sun, Ziyi Yang, Xiaoyang Lyu, Yan-Pei Cao, and Xiaojuan Qi.
\newblock Sc-gs: Sparse-controlled gaussian splatting for editable dynamic scenes.
\newblock In \emph{Proceedings of the IEEE/CVF Conference on Computer Vision and Pattern Recognition}, pages 4220--4230, 2024{\natexlab{b}}.

\bibitem[Katsumata et~al.(2025)Katsumata, Vo, and Nakayama]{katsumata2025compact}
Kai Katsumata, Duc~Minh Vo, and Hideki Nakayama.
\newblock A compact dynamic 3d gaussian representation for real-time dynamic view synthesis.
\newblock In \emph{European Conference on Computer Vision}, pages 394--412. Springer, 2025.

\bibitem[Kerbl et~al.(2023)Kerbl, Kopanas, Leimk{\"u}hler, and Drettakis]{3dgs}
Bernhard Kerbl, Georgios Kopanas, Thomas Leimk{\"u}hler, and George Drettakis.
\newblock 3d gaussian splatting for real-time radiance field rendering.
\newblock \emph{ACM TOG}, 42\penalty0 (4):\penalty0 1--14, 2023.

\bibitem[Labe et~al.(2024)Labe, Issachar, Lang, and Benaim]{labe2024dgd}
Isaac Labe, Noam Issachar, Itai Lang, and Sagie Benaim.
\newblock Dgd: Dynamic 3d gaussians distillation.
\newblock \emph{arXiv preprint arXiv:2405.19321}, 2024.

\bibitem[Lee et~al.(2024)Lee, Won, Jung, Bae, and Jeon]{lee2024fully}
Junoh Lee, Chang-Yeon Won, Hyunjun Jung, Inhwan Bae, and Hae-Gon Jeon.
\newblock Fully explicit dynamic gaussian splatting.
\newblock \emph{arXiv preprint arXiv:2410.15629}, 2024.

\bibitem[Li et~al.(2023{\natexlab{a}})Li, Yu, Choy, Xiao, Alvarez, Fidler, Feng, and Anandkumar]{li2023voxformer}
Yiming Li, Zhiding Yu, Christopher Choy, Chaowei Xiao, Jose~M Alvarez, Sanja Fidler, Chen Feng, and Anima Anandkumar.
\newblock Voxformer: Sparse voxel transformer for camera-based 3d semantic scene completion.
\newblock In \emph{CVPR}, pages 9087--9098, 2023{\natexlab{a}}.

\bibitem[Li et~al.(2022)Li, Wang, Li, Xie, Sima, Lu, Qiao, and Dai]{li2022bevformer}
Zhiqi Li, Wenhai Wang, Hongyang Li, Enze Xie, Chonghao Sima, Tong Lu, Yu Qiao, and Jifeng Dai.
\newblock Bevformer: Learning bird’s-eye-view representation from multi-camera images via spatiotemporal transformers.
\newblock In \emph{ECCV}, pages 1--18. Springer, 2022.

\bibitem[Li et~al.(2023{\natexlab{b}})Li, Yu, Austin, Fang, Lan, Kautz, and Alvarez]{li2023fb}
Zhiqi Li, Zhiding Yu, David Austin, Mingsheng Fang, Shiyi Lan, Jan Kautz, and Jose~M Alvarez.
\newblock Fb-occ: 3d occupancy prediction based on forward-backward view transformation.
\newblock \emph{arXiv preprint arXiv:2307.01492}, 2023{\natexlab{b}}.

\bibitem[Lin et~al.(2024)Lin, Dai, Zhu, and Yao]{lin2024gaussian}
Youtian Lin, Zuozhuo Dai, Siyu Zhu, and Yao Yao.
\newblock Gaussian-flow: 4d reconstruction with dynamic 3d gaussian particle.
\newblock In \emph{Proceedings of the IEEE/CVF Conference on Computer Vision and Pattern Recognition}, pages 21136--21145, 2024.

\bibitem[Liu et~al.(2023)Liu, Wang, Chen, Yang, Zeng, Chen, and Wang]{liu2023fully}
Haisong Liu, Haiguang Wang, Yang Chen, Zetong Yang, Jia Zeng, Li Chen, and Limin Wang.
\newblock Fully sparse 3d panoptic occupancy prediction.
\newblock \emph{arXiv preprint arXiv:2312.17118}, 2023.

\bibitem[Liu et~al.(2024{\natexlab{a}})Liu, Wang, Xie, Liu, Liu, Tian, Yang, and Wang]{liu2024surroundsdf}
Lizhe Liu, Bohua Wang, Hongwei Xie, Daqi Liu, Li Liu, Zhiqiang Tian, Kuiyuan Yang, and Bing Wang.
\newblock Surroundsdf: Implicit 3d scene understanding based on signed distance field.
\newblock \emph{arXiv preprint arXiv:2403.14366}, 2024{\natexlab{a}}.

\bibitem[Liu et~al.(2024{\natexlab{b}})Liu, Liu, Wang, Lv, Wang, Wang, and Hou]{liu2024modgs}
Qingming Liu, Yuan Liu, Jiepeng Wang, Xianqiang Lv, Peng Wang, Wenping Wang, and Junhui Hou.
\newblock Modgs: Dynamic gaussian splatting from causually-captured monocular videos.
\newblock \emph{arXiv preprint arXiv:2406.00434}, 2024{\natexlab{b}}.

\bibitem[Lu et~al.(2024)Lu, Guo, Hui, Chen, Yang, Tang, Zhu, and Dai]{lu20243d}
Zhicheng Lu, Xiang Guo, Le Hui, Tianrui Chen, Min Yang, Xiao Tang, Feng Zhu, and Yuchao Dai.
\newblock 3d geometry-aware deformable gaussian splatting for dynamic view synthesis.
\newblock \emph{arXiv preprint arXiv:2404.06270}, 2024.

\bibitem[Luiten et~al.(2023)Luiten, Kopanas, Leibe, and Ramanan]{luiten2023dynamic}
Jonathon Luiten, Georgios Kopanas, Bastian Leibe, and Deva Ramanan.
\newblock Dynamic 3d gaussians: Tracking by persistent dynamic view synthesis.
\newblock \emph{arXiv preprint arXiv:2308.09713}, 2023.

\bibitem[Lyu et~al.(2024)Lyu, Sun, Huang, Wu, Yang, Chen, Pang, and Qi]{lyu20243dgsr}
Xiaoyang Lyu, Yang-Tian Sun, Yi-Hua Huang, Xiuzhe Wu, Ziyi Yang, Yilun Chen, Jiangmiao Pang, and Xiaojuan Qi.
\newblock 3dgsr: Implicit surface reconstruction with 3d gaussian splatting.
\newblock \emph{arXiv preprint arXiv:2404.00409}, 2024.

\bibitem[Ma et~al.(2024{\natexlab{a}})Ma, Chen, Huang, Xu, Luo, Xu, Gu, Ai, and Wang]{ma2024cam4docc}
Junyi Ma, Xieyuanli Chen, Jiawei Huang, Jingyi Xu, Zhen Luo, Jintao Xu, Weihao Gu, Rui Ai, and Hesheng Wang.
\newblock Cam4docc: Benchmark for camera-only 4d occupancy forecasting in autonomous driving applications.
\newblock In \emph{Proceedings of the IEEE/CVF Conference on Computer Vision and Pattern Recognition}, pages 21486--21495, 2024{\natexlab{a}}.

\bibitem[Ma et~al.(2024{\natexlab{b}})Ma, Tan, Qu, Ma, Zhang, and Xie]{ma2024cotr}
Qihang Ma, Xin Tan, Yanyun Qu, Lizhuang Ma, Zhizhong Zhang, and Yuan Xie.
\newblock Cotr: Compact occupancy transformer for vision-based 3d occupancy prediction.
\newblock In \emph{Proceedings of the IEEE/CVF Conference on Computer Vision and Pattern Recognition}, pages 19936--19945, 2024{\natexlab{b}}.

\bibitem[Miao et~al.(2023)Miao, Liu, Chen, Gong, Xu, Hu, and Zhou]{miao2023occdepth}
Ruihang Miao, Weizhou Liu, Mingrui Chen, Zheng Gong, Weixin Xu, Chen Hu, and Shuchang Zhou.
\newblock Occdepth: A depth-aware method for 3d semantic scene completion.
\newblock \emph{arXiv preprint arXiv:2302.13540}, 2023.

\bibitem[Mildenhall et~al.(2021)Mildenhall, Srinivasan, Tancik, Barron, Ramamoorthi, and Ng]{mildenhall2021nerf}
Ben Mildenhall, Pratul~P Srinivasan, Matthew Tancik, Jonathan~T Barron, Ravi Ramamoorthi, and Ren Ng.
\newblock Nerf: Representing scenes as neural radiance fields for view synthesis.
\newblock \emph{Communications of the ACM}, 65\penalty0 (1):\penalty0 99--106, 2021.

\bibitem[Pan et~al.(2023)Pan, Liu, Zhang, Huang, Li, Liu, and Zhang]{pan2023renderocc}
Mingjie Pan, Jiaming Liu, Renrui Zhang, Peixiang Huang, Xiaoqi Li, Li Liu, and Shanghang Zhang.
\newblock Renderocc: Vision-centric 3d occupancy prediction with 2d rendering supervision.
\newblock \emph{arXiv preprint arXiv:2309.09502}, 2023.

\bibitem[Peng et~al.(2024)Peng, Xu, Cheng, Yang, Wu, Qian, Wang, Wu, and Cai]{peng2024learning}
Liang Peng, Junkai Xu, Haoran Cheng, Zheng Yang, Xiaopei Wu, Wei Qian, Wenxiao Wang, Boxi Wu, and Deng Cai.
\newblock Learning occupancy for monocular 3d object detection.
\newblock In \emph{Proceedings of the IEEE/CVF Conference on Computer Vision and Pattern Recognition}, pages 10281--10292, 2024.

\bibitem[Sima et~al.(2023)Sima, Tong, Wang, Chen, Wu, Deng, Gu, Lu, Luo, Lin, and Li]{sima2023_occnet}
Chonghao Sima, Wenwen Tong, Tai Wang, Li Chen, Silei Wu, Hanming Deng, Yi Gu, Lewei Lu, Ping Luo, Dahua Lin, and Hongyang Li.
\newblock Scene as occupancy.
\newblock \emph{Proceedings of the IEEE/CVF International Conference on Computer Vision}, 2023.

\bibitem[Song et~al.(2024)Song, Liang, Cao, Yan, Zimmer, Gross, Festag, and Knoll]{song2024collaborative}
Rui Song, Chenwei Liang, Hu Cao, Zhiran Yan, Walter Zimmer, Markus Gross, Andreas Festag, and Alois Knoll.
\newblock Collaborative semantic occupancy prediction with hybrid feature fusion in connected automated vehicles.
\newblock In \emph{Proceedings of the IEEE/CVF Conference on Computer Vision and Pattern Recognition}, pages 17996--18006, 2024.

\bibitem[Tang et~al.(2023)Tang, Ren, Zhou, Liu, and Zeng]{tang2023dreamgaussian}
Jiaxiang Tang, Jiawei Ren, Hang Zhou, Ziwei Liu, and Gang Zeng.
\newblock Dreamgaussian: Generative gaussian splatting for efficient 3d content creation.
\newblock \emph{arXiv preprint arXiv:2309.16653}, 2023.

\bibitem[Tang et~al.(2024)Tang, Wang, Wang, Zheng, Ren, Feng, and Ma]{tang2024sparseocc}
Pin Tang, Zhongdao Wang, Guoqing Wang, Jilai Zheng, Xiangxuan Ren, Bailan Feng, and Chao Ma.
\newblock Sparseocc: Rethinking sparse latent representation for vision-based semantic occupancy prediction.
\newblock In \emph{Proceedings of the IEEE/CVF Conference on Computer Vision and Pattern Recognition}, pages 15035--15044, 2024.

\bibitem[Tian et~al.(2024)Tian, Jiang, Yun, Mao, Yang, Wang, Wang, and Zhao]{tian2024occ3d}
Xiaoyu Tian, Tao Jiang, Longfei Yun, Yucheng Mao, Huitong Yang, Yue Wang, Yilun Wang, and Hang Zhao.
\newblock Occ3d: A large-scale 3d occupancy prediction benchmark for autonomous driving.
\newblock \emph{NeurIPS}, 36, 2024.

\bibitem[Tong et~al.(2023)Tong, Sima, Wang, Chen, Wu, Deng, Gu, Lu, Luo, Lin, et~al.]{tong2023scene}
Wenwen Tong, Chonghao Sima, Tai Wang, Li Chen, Silei Wu, Hanming Deng, Yi Gu, Lewei Lu, Ping Luo, Dahua Lin, et~al.
\newblock Scene as occupancy.
\newblock In \emph{ICCV}, pages 8406--8415, 2023.

\bibitem[Vobecky et~al.(2024)Vobecky, Sim{\'e}oni, Hurych, Gidaris, Bursuc, P{\'e}rez, and Sivic]{vobecky2024pop}
Antonin Vobecky, Oriane Sim{\'e}oni, David Hurych, Spyridon Gidaris, Andrei Bursuc, Patrick P{\'e}rez, and Josef Sivic.
\newblock Pop-3d: Open-vocabulary 3d occupancy prediction from images.
\newblock \emph{Advances in Neural Information Processing Systems}, 36, 2024.

\bibitem[Wang et~al.(2023{\natexlab{a}})Wang, Zhu, Xu, Zhang, Wei, Chi, Ye, Du, Lu, and Wang]{wang2023openoccupancy}
Xiaofeng Wang, Zheng Zhu, Wenbo Xu, Yunpeng Zhang, Yi Wei, Xu Chi, Yun Ye, Dalong Du, Jiwen Lu, and Xingang Wang.
\newblock Openoccupancy: A large scale benchmark for surrounding semantic occupancy perception.
\newblock In \emph{ICCV}, pages 17850--17859, 2023{\natexlab{a}}.

\bibitem[Wang et~al.(2023{\natexlab{b}})Wang, Chen, Liao, Fan, and Zhang]{wang2023panoocc}
Yuqi Wang, Yuntao Chen, Xingyu Liao, Lue Fan, and Zhaoxiang Zhang.
\newblock Panoocc: Unified occupancy representation for camera-based 3d panoptic segmentation.
\newblock \emph{arXiv preprint arXiv:2306.10013}, 2023{\natexlab{b}}.

\bibitem[Wei et~al.(2025)Wei, Li, and Liu]{wei2024omniscene}
Dongxu Wei, Zhiqi Li, and Peidong Liu.
\newblock Omni-scene: omni-gaussian representation for ego-centric sparse-view scene reconstruction.
\newblock In \emph{Proceedings of the IEEE/CVF Conference on Computer Vision and Pattern Recognition}, 2025.

\bibitem[Wei et~al.(2023)Wei, Zhao, Zheng, Zhu, Zhou, and Lu]{wei2023surroundocc}
Yi Wei, Linqing Zhao, Wenzhao Zheng, Zheng Zhu, Jie Zhou, and Jiwen Lu.
\newblock Surroundocc: Multi-camera 3d occupancy prediction for autonomous driving.
\newblock In \emph{Proceedings of the IEEE/CVF International Conference on Computer Vision}, pages 21729--21740, 2023.

\bibitem[Wu et~al.(2024)Wu, Yan, Wang, Li, Hui, and Yang]{wu2024deep}
Yuan Wu, Zhiqiang Yan, Zhengxue Wang, Xiang Li, Le Hui, and Jian Yang.
\newblock Deep height decoupling for precise vision-based 3d occupancy prediction.
\newblock \emph{arXiv preprint arXiv:2409.07972}, 2024.

\bibitem[Yan et~al.(2024{\natexlab{a}})Yan, Lin, Zhou, Wang, Sun, Zhan, Lang, Zhou, and Peng]{yan2024street}
Yunzhi Yan, Haotong Lin, Chenxu Zhou, Weijie Wang, Haiyang Sun, Kun Zhan, Xianpeng Lang, Xiaowei Zhou, and Sida Peng.
\newblock Street gaussians for modeling dynamic urban scenes.
\newblock \emph{arXiv preprint arXiv:2401.01339}, 2024{\natexlab{a}}.

\bibitem[Yan et~al.(2022)Yan, Wang, Li, Zhang, Li, and Yang]{yan2022rignet}
Zhiqiang Yan, Kun Wang, Xiang Li, Zhenyu Zhang, Jun Li, and Jian Yang.
\newblock Rignet: Repetitive image guided network for depth completion.
\newblock In \emph{European Conference on Computer Vision}, pages 214--230. Springer, 2022.

\bibitem[Yan et~al.(2024{\natexlab{b}})Yan, Lin, Wang, Zheng, Wang, Zhang, Li, and Yang]{yan2024tri}
Zhiqiang Yan, Yuankai Lin, Kun Wang, Yupeng Zheng, Yufei Wang, Zhenyu Zhang, Jun Li, and Jian Yang.
\newblock Tri-perspective view decomposition for geometry-aware depth completion.
\newblock In \emph{Proceedings of the IEEE/CVF Conference on Computer Vision and Pattern Recognition}, pages 4874--4884, 2024{\natexlab{b}}.

\bibitem[Yang et~al.(2023)Yang, Gao, Zhou, Jiao, Zhang, and Jin]{yang2023deformable}
Ziyi Yang, Xinyu Gao, Wen Zhou, Shaohui Jiao, Yuqing Zhang, and Xiaogang Jin.
\newblock Deformable 3d gaussians for high-fidelity monocular dynamic scene reconstruction.
\newblock \emph{arXiv preprint arXiv:2309.13101}, 2023.

\bibitem[Yi et~al.(2023)Yi, Fang, Wu, Xie, Zhang, Liu, Tian, and Wang]{yi2023gaussiandreamer}
Taoran Yi, Jiemin Fang, Guanjun Wu, Lingxi Xie, Xiaopeng Zhang, Wenyu Liu, Qi Tian, and Xinggang Wang.
\newblock Gaussiandreamer: Fast generation from text to 3d gaussian splatting with point cloud priors.
\newblock \emph{arXiv preprint arXiv:2310.08529}, 2023.

\bibitem[Yu et~al.(2023)Yu, Shu, Deng, Lu, Liu, Yu, Yang, Li, and Chen]{yu2023flashocc}
Zichen Yu, Changyong Shu, Jiajun Deng, Kangjie Lu, Zongdai Liu, Jiangyong Yu, Dawei Yang, Hui Li, and Yan Chen.
\newblock Flashocc: Fast and memory-efficient occupancy prediction via channel-to-height plugin.
\newblock \emph{arXiv preprint arXiv:2311.12058}, 2023.

\bibitem[Zhang et~al.(2023{\natexlab{a}})Zhang, Yan, Wei, Li, Liu, Tang, Duan, and Lu]{zhang2023occnerf}
Chubin Zhang, Juncheng Yan, Yi Wei, Jiaxin Li, Li Liu, Yansong Tang, Yueqi Duan, and Jiwen Lu.
\newblock Occnerf: Self-supervised multi-camera occupancy prediction with neural radiance fields.
\newblock \emph{arXiv e-prints}, pages arXiv--2312, 2023{\natexlab{a}}.

\bibitem[Zhang et~al.(2024)Zhang, Li, Li, Bressieux, Hilliges, Pollefeys, Van~Gool, and Wang]{zhang2024egogaussian}
Daiwei Zhang, Gengyan Li, Jiajie Li, Micka{\"e}l Bressieux, Otmar Hilliges, Marc Pollefeys, Luc Van~Gool, and Xi Wang.
\newblock Egogaussian: Dynamic scene understanding from egocentric video with 3d gaussian splatting.
\newblock \emph{arXiv preprint arXiv:2406.19811}, 2024.

\bibitem[Zhang et~al.(2023{\natexlab{b}})Zhang, Zhu, and Du]{zhang2023occformer}
Yunpeng Zhang, Zheng Zhu, and Dalong Du.
\newblock Occformer: Dual-path transformer for vision-based 3d semantic occupancy prediction.
\newblock In \emph{ICCV}, pages 9433--9443, 2023{\natexlab{b}}.

\bibitem[Zhao et~al.(2024{\natexlab{a}})Zhao, Sun, Wang, Guo, Wan, Huang, Huang, Chen, and Ren]{zhao2024tclc}
Cheng Zhao, Su Sun, Ruoyu Wang, Yuliang Guo, Jun-Jun Wan, Zhou Huang, Xinyu Huang, Yingjie~Victor Chen, and Liu Ren.
\newblock Tclc-gs: Tightly coupled lidar-camera gaussian splatting for surrounding autonomous driving scenes.
\newblock \emph{arXiv preprint arXiv:2404.02410}, 2024{\natexlab{a}}.

\bibitem[Zhao et~al.(2024{\natexlab{b}})Zhao, Xu, Wang, Zhang, Zhang, Zheng, Du, Zhou, and Lu]{zhao2024lowrankocc}
Linqing Zhao, Xiuwei Xu, Ziwei Wang, Yunpeng Zhang, Borui Zhang, Wenzhao Zheng, Dalong Du, Jie Zhou, and Jiwen Lu.
\newblock Lowrankocc: Tensor decomposition and low-rank recovery for vision-based 3d semantic occupancy prediction.
\newblock In \emph{Proceedings of the IEEE/CVF Conference on Computer Vision and Pattern Recognition}, pages 9806--9815, 2024{\natexlab{b}}.

\bibitem[Zheng et~al.(2025)Zheng, Chen, Huang, Zhang, Duan, and Lu]{zheng2025occworld}
Wenzhao Zheng, Weiliang Chen, Yuanhui Huang, Borui Zhang, Yueqi Duan, and Jiwen Lu.
\newblock Occworld: Learning a 3d occupancy world model for autonomous driving.
\newblock In \emph{European Conference on Computer Vision}, pages 55--72. Springer, 2025.

\bibitem[Zhu et~al.(2024)Zhu, Liang, Chang, Deng, Lu, Yang, Zhang, and Zhang]{zhu2024motiongs}
Ruijie Zhu, Yanzhe Liang, Hanzhi Chang, Jiacheng Deng, Jiahao Lu, Wenfei Yang, Tianzhu Zhang, and Yongdong Zhang.
\newblock Motiongs: Exploring explicit motion guidance for deformable 3d gaussian splatting.
\newblock \emph{arXiv preprint arXiv:2410.07707}, 2024.

\end{thebibliography}
}


\end{document}